\documentclass[10pt,journal,compsoc]{IEEEtran}
\ifCLASSOPTIONcompsoc
  \usepackage[nocompress]{cite}
\else
  \usepackage{cite}
\fi
\ifCLASSINFOpdf
\else
\fi
\usepackage{amssymb}
\usepackage{hyperref}
\usepackage{url}            
\usepackage{booktabs}       
\usepackage{amsfonts}       
\usepackage{nicefrac}       
\usepackage{microtype}      
\usepackage{xcolor}         
\usepackage{colortbl}
\usepackage{wrapfig}
\usepackage{graphicx}
\usepackage{lipsum}
\usepackage{multirow}
\usepackage{textcomp}
\usepackage{pifont} 
\usepackage{caption}
\usepackage{subcaption}
\newcommand{\cmark}{\ding{51}}%
\newcommand{\xmark}{\ding{55}}%

\newcommand{\nameShort}{LAUDNet}

\begin{document}
%
\title{Latency-aware Unified Dynamic Networks for Efficient Image Recognition}

%
%

\author{Yizeng~Han\IEEEauthorrefmark{1},
        Zeyu~Liu\IEEEauthorrefmark{1},
        Zhihang~Yuan\IEEEauthorrefmark{1},
        Yifan~Pu,~Chaofei Wang\\
        Shiji~Song,~\IEEEmembership{Senior~Member,~IEEE,}        and~Gao~Huang~\IEEEmembership{Member,~IEEE}
\IEEEcompsocitemizethanks{\IEEEcompsocthanksitem Yizeng Han, Yifan Pu, Chaofei Wang, Shiji Song, and Gao Huang are with the  Department
of Automation, Tsinghua University. Gao Huang is also with Beijing Academy
of Artificial Intelligence.  
E-mail: \{hanyz18, pyf20, wangcf18\}@mails.tsinghua.edu.cn, \{gaohuang, shijis\}@tsinghua.edu.cn. Corresponding author: Gao Huang.
\IEEEcompsocthanksitem Zeyu Liu is with the Department of Computer Science and Technology, Tsinghua University. E-mail: liuzeyu20@mails.tsinghua.edu.cn.
\IEEEcompsocthanksitem Zhihang Yuan is with Houmo.AI. Email: hahnyuan@gmail.com.
}
}

\IEEEtitleabstractindextext{%
\begin{abstract}
    Dynamic computation has emerged as a promising strategy to improve the inference efficiency of deep networks. 
    It allows selective activation of various computing units, such as layers or convolution channels, or adaptive allocation of computation to highly informative spatial regions in image features, thus significantly reducing unnecessary computations conditioned on each input sample. 
    However, the practical efficiency of dynamic models does not always correspond to theoretical outcomes. This discrepancy stems from three key challenges: 1) The absence of a \emph{unified formulation} for various dynamic inference paradigms, owing to the fragmented research landscape; 2) The undue emphasis on algorithm design while neglecting \emph{scheduling strategies}, which are critical for optimizing computational performance and resource utilization in CUDA-enabled GPU settings; and 3) The cumbersome process of evaluating practical latency, as most existing libraries are tailored for static operators. To address these issues, we introduce \textbf{Latency-Aware Unified Dynamic Networks (LAUDNet)}, a comprehensive framework that amalgamates three cornerstone dynamic paradigms—spatially-adaptive computation, dynamic layer skipping, and dynamic channel skipping—under a unified formulation. To reconcile theoretical and practical efficiency, LAUDNet integrates algorithmic design with scheduling optimization, assisted by a latency predictor that accurately and efficiently gauges the inference latency of dynamic operators. This latency predictor harmonizes considerations of algorithms, scheduling strategies, and hardware attributes. We empirically validate various dynamic paradigms within the LAUDNet framework across a range of vision tasks, including image classification, object detection, and instance segmentation. Our experiments confirm that LAUDNet effectively narrows the gap between theoretical and real-world efficiency. For example, LAUDNet can reduce the practical latency of its static counterpart, ResNet-101, by over 50\% on hardware platforms such as V100, RTX3090, and TX2 GPUs. Furthermore, LAUDNet surpasses competing methods in the trade-off between accuracy and efficiency. Code is available at: \url{https://www.github.com/LeapLabTHU/LAUDNet}.
\end{abstract}

}
\maketitle


\begingroup\renewcommand\thefootnote{\IEEEauthorrefmark{1}}
\footnotetext{\emph{Equal contribution.}}
\endgroup

\IEEEdisplaynontitleabstractindextext

%
\IEEEpeerreviewmaketitle

\IEEEraisesectionheading{\section{Introduction}\label{sec:introduction}}

\begin{figure*}
\centering
\includegraphics[width=0.95\textwidth]{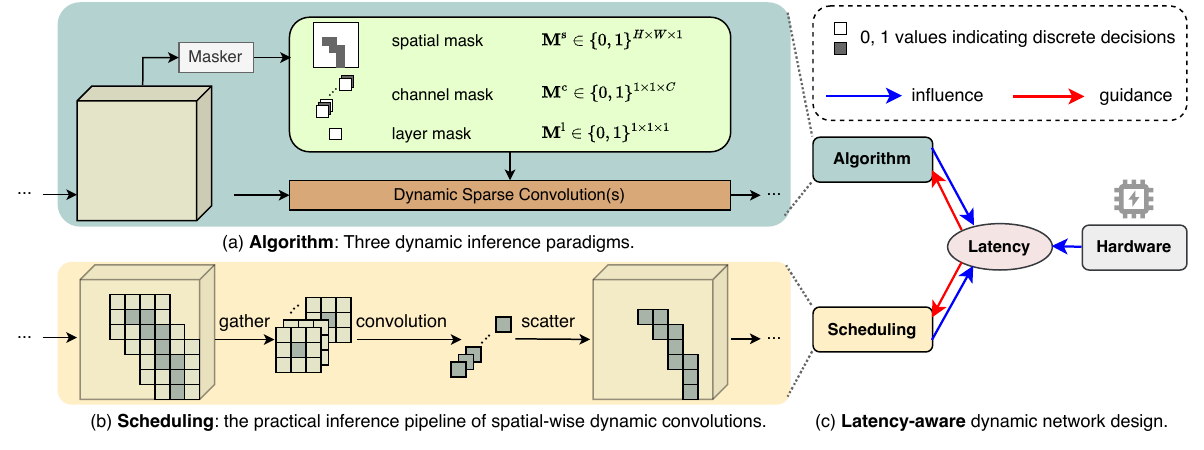}
\vskip -0.2in
\caption{An overview of our method. (a) illustrates three representative adaptive inference \emph{algorithms} (\emph{i.e.} spatial-wise dynamic convolution, channel skipping, and layer skipping); (b) is an example of the \emph{scheduling} strategy for spatial-wise dynamic convolution; and (c) presents our key idea of using the latency to \emph{guide} both algorithm design and scheduling optimization.}
\vskip -0.2in
\label{fig:overview}
\end{figure*}

\IEEEPARstart{D}{eep} neural networks have demonstrated exceptional capabilities in various domains such as computer vision \cite{deng2009imagenet,he2016resnet,huang2017densely,dosovitskiy2020image,kirillov2023segment}, natural language processing \cite{vaswani2017attention,devlin_bert_2019,radford2018improving,radford2019language}, and multi-modal understanding/generation \cite{openai2023gpt4}. Despite their stellar performance, the intensive computational requirements of these deep networks often limit their deployment on resource-constrained platforms, like mobile phones and IoT devices, highlighting the need for more efficient deep learning models.

Unlike traditional static networks \cite{he2016resnet,huang2017densely,dosovitskiy2020image} which process all inputs uniformly, dynamic models \cite{han2021dynamic} adaptively allocate computation in a data-dependent fashion. This adaptivity involves bypassing certain network layers \cite{huang2017multi,han2022learning,wang2018skipnet,veit2018convolutional} or convolution channels \cite{lin2017runtime,bejnordi2019batch} conditionally, and executing spatially adaptive inference that concentrates computational effort on the most informative regions of an image \cite{figurnov2017spatially,dong_more_2017,verelst_dynamic_2020,xie2020spatially,wang2020glance,SAR_TIP}. As the field evolves and various dynamic models show promise, it begs the question: \emph{How can we design a dynamic network for practical use?}

Addressing this question is challenging due to difficulties in fairly comparing different dynamic-computation paradigms. These challenges fall into three categories: 1) The lack of a unified framework to encompass different paradigms, as research in this area is often fragmented; 2) The focus on algorithm design, which often results in the mismatch between practical efficiency and their theoretical computational potential, due to the significant impact of scheduling strategies\footnote{Scheduling strategies are essential for practical efficiency because they optimize the use of GPU threads and memory with CUDA codes.} and hardware properties on real-world latency; 3) The laborious task of evaluating a dynamic model's latency on different hardware platforms, as common libraries (\emph{e.g.} cuDNN) are not built to accelerate many dynamic operators.

In response, we introduce a \textbf{Latency-Aware Unified Dynamic Network (\nameShort)}, a framework that unifies three representative dynamic-inference paradigms. Specifically, we examine the algorithmic design of layer skipping, channel skipping, and spatially dynamic convolution, integrating them through a "mask-and-compute" scheme (\figurename~\ref{fig:overview} (a)).


Next, we delve into the challenges of translating theoretical efficiency into tangible speedup, especially on multi-core processors such as GPUs. Traditional literature commonly adopts hardware-agnostic FLOPs (floating-point operations) as a crude efficiency measure, failing to provide latency-aware guidance for algorithm design. In dynamic networks, adaptive computation coupled with sub-optimal scheduling strategies intensifies the gap between FLOPs and latency. Moreover, most existing methods execute adaptive inference at the finest granularity. For instance, in spatial-wise dynamic inference, the decision to compute each feature pixel is made independently \cite{dong_more_2017,verelst_dynamic_2020,xie2020spatially}. This fine-grained flexibility results in non-contiguous memory access \cite{xie2020spatially}, necessitating specialized scheduling strategies (\figurename~\ref{fig:overview} (b)).

Given that dynamic operators exhibit unique memory access patterns and scheduling strategies, libraries designed for static models, like cuDNN, fail to optimize dynamic models effectively. The lack of library support implies that each dynamic operator requires individualized scheduling optimization, code refinement, compilation, and deployment, making network latency evaluation across hardware platforms labor-intensive. To address this, we propose a novel latency prediction model that efficiently estimates network latency by taking into account algorithm design, scheduling strategies, and hardware properties. Compared to hardware-agnostic FLOPs, our predicted latency offers a more realistic representation of dynamic model efficiency.

Guided by the latency prediction model, we tackle the aforementioned challenges within our latency-aware unified dynamic network (\nameShort) framework. For a given hardware device, we use the predicted latency as the guiding metric for algorithm design and scheduling optimization, as opposed to the conventionally used FLOPs (\figurename~\ref{fig:overview} (c)). In this context, we propose coarse-grained dynamic networks where "whether-to-compute" decisions are made at the patch/group level rather than individual pixels/channels. Though less flexible than pixel/channel-level adaptability in prior works \cite{dong_more_2017,verelst_dynamic_2020,xie2020spatially,lin2017runtime,bejnordi2019batch}, this approach encourages contiguous memory access, enhancing real-world speedup on hardware. Our improved scheduling strategies further permit batching inference. We investigate dynamic inference paradigms, focusing on the accuracy-latency trade-off. Notably, previous research has established a correlation between latency and FLOPs on CPUs \cite{SAR_TIP,xie2020spatially}, hence in this paper, we primarily target the GPU platform, a more challenging but less explored environment.

The \nameShort~is designed as a general framework in two ways:
1) \textcolor{black}{Multiple adaptive inference paradigms can be easily implemented in various vision backbones, like ResNets \cite{he2016resnet}, RegNets \cite{radosavovic2020designing} and vision Transformers \cite{touvron2021training,yuan2021tokens}}; and
2) The latency predictor functions as an off-the-shelf tool that can be readily applied to diverse computing platforms, such as server-end GPUs (Tesla V100, RTX3090), desktop-level GPU (RTX3060) and edge devices (Jetson TX2, Nvidia Nano).

We evaluate \nameShort's performance across multiple backbones for image classification, object detection, and instance segmentation. Our results show that \nameShort~significantly improves the efficiency of deep CNNs, both in theory and practice. For instance, the inference latency of ResNet-101 on ImageNet \cite{deng2009imagenet} is reduced by $>$50\% on different types of GPUs (\emph{e.g.}, V100, RTX3090 and TX2), without compromising accuracy. Moreover, our method outperforms various lightweight networks in low-FLOPs scenarios.

Although parts of this work were initially published in a conference version \cite{han2022latency}, this paper significantly expands our previous efforts in several key areas:
\begin{itemize}
    \item A unified dynamic-inference framework is proposed. While the preliminary paper \cite{han2022latency} predominantly focused on spatially adaptive computation, this paper delves deeper into two additional and important dynamic paradigms, specifically, dynamic layer skipping and channel skipping (\figurename~\ref{fig:overview} and Sec.\ref{sec:preliminary}). Furthermore, we integrate these paradigms into a unified framework, and provide more thorough study on architecture design and complexity analysis (Sec.\ref{sec:arch}).
    \item The latency predictor has been enhanced to support an expanded set of dynamic operators, including layer skipping and channel skipping (Sec.~\ref{sec_predictor}). Moreover, we adopt Nvidia Cutlass \cite{cutlass} to optimize the scheduling strategies. Hardware evaluations demonstrate that our latency predictor can accurately predict the latency on real hardware (\figurename~\ref{real_predicted_latency}).
    \item \textcolor{black}{The LAUDNet framework has been extended to accommodate Transformer architectures, as detailed in Sec.~\ref{sec:arch}. This extension notably enhances latency optimization through the implementation of dynamic \emph{token skipping} (spatially adaptive computation), \emph{head (channel) skipping}, and \emph{layer skipping}. Such advancements significantly broaden the applicability of LAUDNet. The empirical evaluation, illustrated in \figurename~\ref{fig:main_results} (c) and discussed in Sec.~\ref{sec:main_results}, yields valuable insights into the design of efficient Transformers, underpinning the framework's versatility and efficacy.}
    \item For the first time, we incorporate batching inference for our dynamic operators (Sec.~\ref{sec_schedule_optim}). This innovation leads to more consistent prediction outcomes and an enhanced speedup ratio on GPU platforms (\figurename~\ref{fig:latency_bs},~\ref{fig:latency_bs_3060}).
    \item We undertake an exhaustive analysis of various dynamic granularities (\figurename~\ref{fig:compare_S_G}) and paradigms (\figurename~\ref{fig:main_results},\ref{fig_vis},\ref{fig_3060_nano}, Tab.~\ref{tab:coco},\ref{tab:coco_seg}), spanning different vision tasks and platforms, with added evaluations on contemporary GPUs like RTX3060 and RTX3090. We are confident that our results will offer valuable insights to both researchers and practitioners.    
\end{itemize}

\section{Related works}

\noindent\textbf{Efficient deep learning} has garnered substantial interest. Traditional solutions involve lightweight model design \cite{howard2017mobilenets,sandler2018mobilenetv2,zhang2018shufflenet,ma2018shufflenet}, network pruning \cite{han2015deepcompression,he2018pruning,huang2018condensenet,yang2021condensenet}, weight quantization \cite{hubara2016binarized,choi2018pact,jung2019learning,yuan2022ptq4vit}, and knowledge distillation \cite{hinton2014distilling,wang2022efficient}. However, these \emph{static} methods have sub-optimal inference strategy, leading to intrinsic redundancy since they process all inputs with equal computation.

\noindent\textbf{Dynamic networks} \cite{han2021dynamic,graves2016adaptive,figurnov2017spatially,huang2017multi} propose an appealing alternative to static models by enabling input-conditional \emph{dynamic inference}. This adaptive approach has yielded superior results across various domains. In visual recognition, prevalent dynamic paradigms include early exiting \cite{huang2017multi,yang2020resolution,han2022learning}, layer skipping \cite{graves2016adaptive,wang2018skipnet,veit2018convolutional}, channel skipping \cite{lin2017runtime,gao2018dynamic,bejnordi2019batch}, and spatial-wise dynamic computation \cite{dong_more_2017,verelst_dynamic_2020,xie2020spatially}. This paper primarily targets the latter three paradigms, as they can be readily applied to arbitrary visual backbones, thereby offering a generality advantage. Layer skipping and channel skipping explore \emph{structural} redundancy within deep networks by selectively activating computation units, such as layers or convolution channels when processing different inputs. Spatial-wise dynamic models alleviate spatial redundancy in image features and selectively assign computation to the regions most pertinent to the task at hand.

Despite their effectiveness, previous studies often fail to recognize the shared underlying formulation across different dynamic paradigms. In contrast, we introduce a unified framework that encompasses all three paradigms, facilitating a thorough exploration of dynamic networks. Additionally, existing methods primarily concentrate on algorithm design, which often results in a significant disparity between theoretical and practical efficiency. In our latency-aware co-design framework, we bridge this gap by utilizing latency directly from our latency predictor to guide both algorithm design and scheduling optimization. This approach results in improved latency performance across diverse platforms.

\noindent\textbf{Hardware-aware network design.} Researchers have acknowledged the necessity to bridge the gap between theoretical and practical efficiency of deep models by considering actual latency during network design. Two primary approaches have emerged: the first entails conducting speed tests on hardware and deriving guidelines to facilitate \emph{hand-designing} lightweight models \cite{ma2018shufflenet}, and the second involves performing speed tests for various types of \emph{static} operators and modeling the latency predictor as a small trainable model \cite{cai2019proxylessnas,tan2019mnasnet,wu2019fbnet,cai2019once}. Neural architecture search (NAS) techniques \cite{zoph2016neural,liu_darts_2018} are then used to \emph{search} for hardware-friendly models. 

Our work distinguishes itself from these approaches in two significant ways:
1) while existing works predominantly focus on constructing \emph{static} models that inherently exhibit computational redundancy by treating all inputs uniformly, our goal is to design latency-aware \emph{dynamic} models that adjust their computation based on inputs; 
2) conducting speed tests for dynamic operators across various hardware devices can be laborious and impractical. To circumvent this, we propose a latency prediction model that efficiently estimates the inference latency of dynamic operators on any given computing platform. This model accounts for algorithm design, scheduling strategies, and hardware properties simultaneously, providing valuable insights without the need for extensive speed testing.

\begin{figure*}
\centering
\vskip -0.1in
\includegraphics[width=\textwidth]{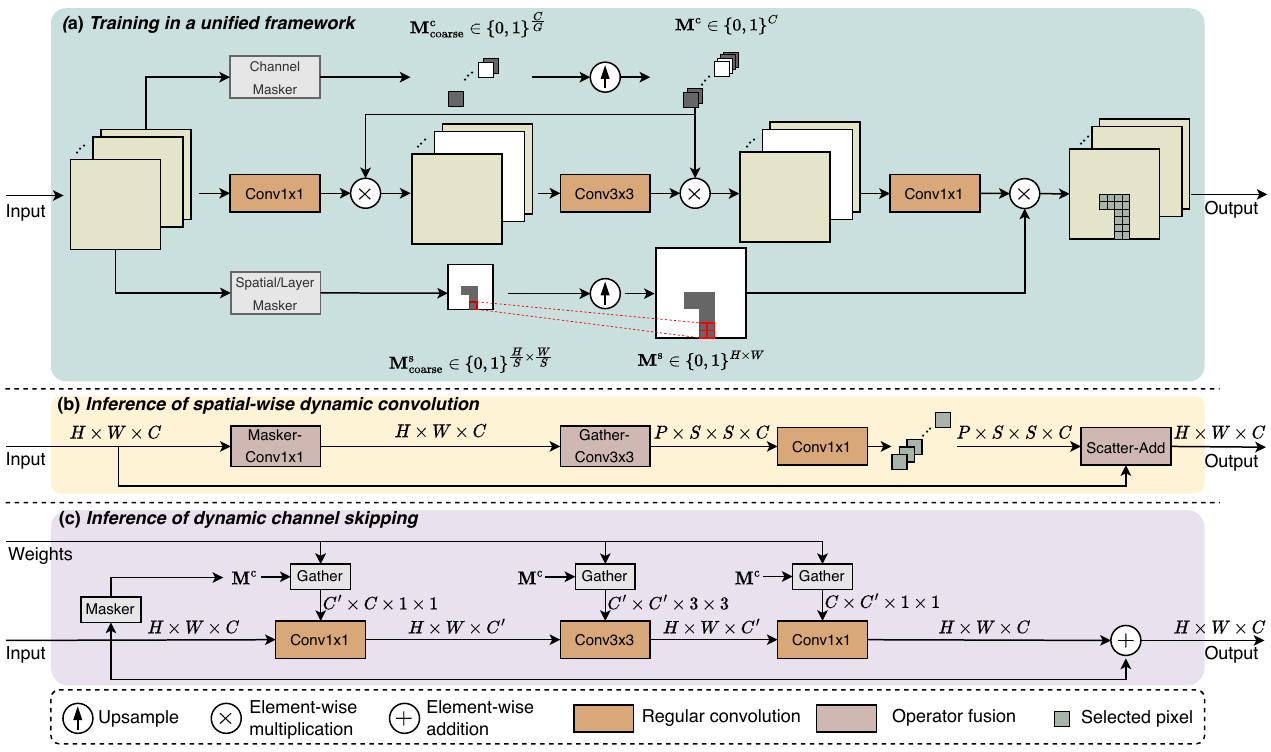}
\vskip -0.1in
\caption{Our proposed \nameShort~block. (a) we first use a lightweight module to generate the channel mask $\mathbf{M}^\mathrm{c}$ or the spatial/layer mask $\mathbf{M}^\mathrm{s}$/$\mathbf{M}^\mathrm{l}$. The granularity of dynamic inference is controlled by $G$ (for channel skipping) and $S$ (for spatially adaptive computation). During training, the channel mask is multiplied with the input and output of the $3\times 3$ convolution, and the spatial mask is applied on the final output of the block. Layer skipping could be easily implemented by setting $S$ equal to the feature resolution. The scheduling strategies in inference ((b) for spatial-wise dynamic convolution and (c) for channel skipping) is performed to decrease memory access and facilitate parallel computation (Sec.~\ref{sec_schedule_optim}). Note that we omit layer skipping here due to its simplicity: the whole block will be executed if the layer masker produces a value of 1.}
\vskip -0.2in
\label{fig:arch}
\end{figure*}

\section{Method}
This section begins by providing an introduction to the foundational concepts underlying three dynamic inference paradigms (Sec.~\ref{sec:preliminary}). We then present the architecture design of our \nameShort~framework, which unifies these paradigms under a cohesive \emph{mask-and-compute} formulation (Sec.~\ref{sec:arch}). Next, we explain the latency prediction model (Sec.~\ref{sec_predictor}), which guides the determination of granularity settings and scheduling optimization (Sec.~\ref{sec_schedule_optim}). Finally, we describe the training strategies for our \nameShort~(Sec.~\ref{sec:train}).
\subsection{Preliminaries}\label{sec:preliminary}
\noindent\textbf{Spatially adaptive computation.} Existing spatial-wise dynamic networks typically incorporate a masker $\mathcal{M}^\mathrm{s}$ within each convolutional block of a CNN backbone. Given an input $\mathbf{x}\!\in\!\mathbb{R}^{H\times W\times C}$ to a block, where $H$ and $W$ represent the feature height and width, and $C$ denotes the channel number. Assuming a convolution stride of 1, the masker $\mathcal{M}^\mathrm{s}$ takes $\mathbf{x}$ as input and generates a binary-valued spatial mask $\mathbf{M}^\mathrm{s}\!=\!\mathcal{M}^\mathrm{s}(\mathbf{x})\!\in\!\left\{0,1\right\}^{H\times W}$. Each element in $\mathbf{M}^\mathrm{s}$ determines whether to perform convolution operations at the corresponding output location. Unselected regions are populated with values from skip connection \cite{dong_more_2017,verelst_dynamic_2020}.

During inference, the current scheduling strategy for spatial-wise dynamic convolutions generally involve three steps \cite{ren_sbnet_2018} (\figurename~\ref{fig:overview} (b)): 1) \emph{gathering}, which re-organizes the selected pixels (if the convolution kernel size is greater than $1\times 1$, the neighbors are also required) along the \emph{batch} dimension; 2) \emph{computation}, which performs convolution on the gathered input; and 3) \emph{scattering}, which fills the computed pixels on their corresponding locations of the output feature. Compared to performing convolutions on the entire feature map, this scheduling strategy reduces computation at the cost of overhead from mask generation and non-contiguous \emph{memory access}. As a result, the overall latency could even be increased, particularly when the \emph{granularity} of dynamic convolution is at the pixel level (\figurename~\ref{fig:r_l_vs_r_s_and_S}).

\noindent\textbf{Dynamic layer skipping} \cite{wu2018blockdrop,wang2018skipnet,veit2018convolutional} adaptively determines whether to execute each layer or block, leveraging the structural redundancy of deep models to achieve data-dependent network \emph{depth}. The implementation of dynamic layer skipping is similar to spatially adaptive inference, but with a scalar ${0,1}$ decision variable $\mathbf{M}^\mathrm{l}$ instead of a spatial $H\times W$ mask. Compared to spatially adaptive inference, layer skipping provides less flexibility but more regular computation patterns. Moreover, it generally does not require special scheduling strategies, as the original convolution operators remain unmodified.

\noindent\textbf{Dynamic channel skipping} \cite{lin2017runtime,bejnordi2019batch,herrmann2018end} takes a more conservative approach to dynamic architecture versus full layer skipping. 
It uses a $C$-dimensional vector $\mathbf{M}^\mathrm{c}\in\{0,1\}^C$ to adaptively determine the runtime \emph{width} of a convolution layer with $C$ output channels. For instance, the $i$-th ($1\le i \le C$) channel is computed only if $\mathbf{M}^\mathrm{c}_i=1$. The scheduling of dynamic channel skipping usually requires gathering convolution kernels instead of feature pixels as in spatially dynamic computation (compare \figurename~\ref{fig:arch} (b) and (c)) .

\subsection{\nameShort~architecture}\label{sec:arch}
\noindent\textbf{Overview}. Our analysis in Sec.\ref{sec:preliminary} reveals that the three dynamic inference paradigms share a common "\emph{mask-and-compute}" scheme, with the key difference being the \emph{mask shapes}. Leveraging this insight, we propose a unified framework (\figurename\ref{fig:arch}) where lightweight modules generate the channel mask $\mathbf{M}^\mathrm{c}$ and the spatial/layer mask $\mathbf{M}^{\mathrm{s}/\mathrm{l}}$, respectively. Notably, layer skipping can be treated as a special case of spatially adaptive inference by introducing the concept of \emph{granularity} in dynamic computation as follows.

\begin{figure} 
\centering 
    \includegraphics[width=0.95\linewidth]{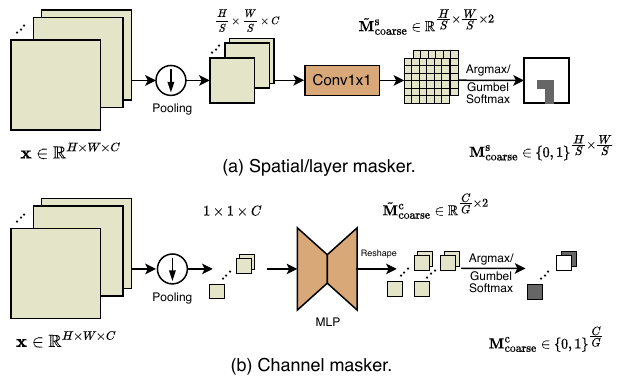} 
    \vskip -0.1in
    \caption{The architecture design of two types of maskers. The spatial/layer masker (a) is composed of a an adaptive pooling layer and a $1\times 1$ convolution. The channel makser (b) consists of a global average pooling and a 2-layer MLP. The argmax operation is directly applied to obtain the discrete decisions during inference, while Gumbel Softmax \cite{jang2016categorical,maddison2016concrete} is utilized for end-to-end training (Sec.~\ref{sec:train}).} 
    \label{fig:maskers}
    \vskip -0.25in
\end{figure}

\noindent\textbf{Dynamic granularity.} As mentioned in Sec.~\ref{sec:preliminary}, using \emph{pixel}-level dynamic convolutions \cite{dong_more_2017,verelst_dynamic_2020,xie2020spatially} poses substantial challenges for realistic speedup on multi-core processors due to non-contiguous memory access. To address this, we propose to optimize the \emph{granularity} of dynamic computation. For spatially adaptive inference, instead of producing an $H\!\times\! W$ mask directly, we first generate a low-resolution mask $\mathbf{M}^\mathrm{s}_{\mathrm{coarse}}\!\in\!\{0,1\}^{\frac{H}{S}\times\frac{W}{S}}$, where $S$ is the \emph{spatial granularity}. Each element in $\mathbf{M}^\mathrm{s}_{\mathrm{coarse}}$ determines computation for a corresponding $S\!\times \!S$ feature patch. For instance, the first ResNet stage\footnote{Here we refer to a stage as the cascading of multiple blocks which process features with the same resolution.} deal with $56\!\times\! 56$ features. Then the valid choices for $S$ are $\left\{1,2,4,7,8,14,28,56\right\}.$ The mask $\mathbf{M}^\mathrm{s}_{\mathrm{coarse}}$ is then upsampled to the size of $H\!\times\! W$. Notably, $S\!=\!1$ corresponds to pixel-level granularity \cite{dong_more_2017,verelst_dynamic_2020,xie2020spatially}, while $S\!=\!56$ naturally implements layer skipping. 

Similarly, we introduce channel granularity $G$ for channel skipping. Each element in $\mathbf{M}^\mathrm{c}_{\mathrm{coarse}}\in\{0,1\}^{\frac{C}{G}}$ determines computation for $G$ feature channels. The choice of the spatial granularity $S$ and the channel granularity $G$ for each block will be guided by our latency predictor (Sec.~\ref{sec_predictor}) for balancing flexibility and efficiency. Note that we apply the channel mask at the first two convolution layers within a block. This design is compatible with various backbone architectures, including those with arbitrary bottleneck ratios or group convolutions \cite{radosavovic2020designing}.


\noindent\textbf{Masker design}. We design different structures for spatial (layer) and channel-wise dynamic computation. As shown in \figurename~\ref{fig:maskers} (a), the spatial masker uses an adaptive pooling layer to downsample the input $\mathbf{x}$ to the size of $\frac{H}{S}\!\times\!\frac{W}{S}\!\times \!C$, followed by a $1\!\times\! 1$ convolution layer producing the soft logits $\tilde{\mathbf{M}}_\mathrm{coarse}^\mathrm{s}\!\in\!\mathbb{R}^{\frac{H}{S}\!\times\!\frac{W}{S}\!\times\! 2}$. For the channel masker, we use a 2-layer MLP (\figurename~\ref{fig:maskers} (b)) to produce channel-skipping decisions. Given input channels $C$ the target mask dimension $D\!=\!C/G$, we set the hidden units in the MLP as $\max\{\lfloor D/16\rfloor, 16\}$, where $\lfloor\cdot\rfloor$ denotes a round-down operation. Appendix~\ref{supp_results_latency_pred} shows this design effectively reduces the latency of channel maskers, especially in late stages with more channels.

\noindent\textbf{Computational complexity.}
We first point out that the masker FLOPs are negligible compared to the backbone convolutions. Therefore, we mainly analyse the complexity of standard convolution blocks here.

For spatially adaptive computation, We define the \emph{activation ratio} $r^\mathrm{s}\!=\!\frac{\sum_{i,j} \mathbf{M}^\mathrm{s}_{i,j}}{H\times W}\!\in\![0,1]$ to denote the fraction of computed pixels. Following \cite{verelst_dynamic_2020}, we further compute $r^\mathrm{s}_\mathrm{dil}$ of a dilated spatial mask to represent the activation ratio of the first convolution in a block. It is observed in our experiments that $r^\mathrm{s}_\mathrm{dil}$ is generally close to $r^\mathrm{s}$. With FLOPs $F_1, F_2, F_3$ for the three convolution layers, the \emph{theoretial} speedup is $\frac{r^\mathrm{s}_\mathrm{dil} F_1 + r^\mathrm{s} F_2 + r^\mathrm{s} F_3}{F_1 + F_2 + F_3} \!\approx \!r^\mathrm{s}$.

\begin{figure}
    \centering
    \includegraphics[width=0.95\linewidth]{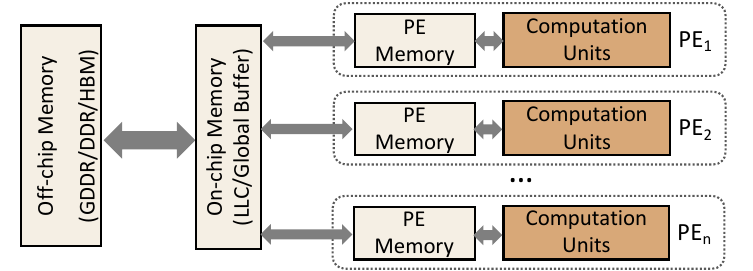} 
    \vskip -0.1in
    \caption{\textcolor{black}{Our hardware model, which allows us to model the latency of both data moving and computation.}} 
    \label{hardware_model} 
    \vskip -0.25in
\end{figure}
For channel skipping, the \emph{activation ratio} is $r^\mathrm{c}=\frac{\sum_j \mathbf{M}^\mathrm{c}_i}{C}\in[0,1]$. Apply the mask before and after the $3\times 3$ convolution makes its complexity quadratic with respect to $r^\mathrm{c}$. The overall speedup is $\frac{r^\mathrm{c} F_1 + (r^\mathrm{c})^2 F_2 + r^\mathrm{c} F_3}{F_1 + F_2 + F_3} \le r^\mathrm{c}$.

\noindent\textcolor{black}{\textbf{Generalization in Transformer architectures.} 
It is essential to highlight that the implementation of the three dynamic paradigms—namely spatial-wise adaptive computation, dynamic channel selection, and layer skipping—is inherently more straightforward in vision Transformers compared to CNNs. These paradigms are not only more amenable to hardware considerations, requiring minimal scheduling optimization, but also benefit from the inherent structure of vision Transformers. For instance, spatial-wise dynamic computation can be efficiently executed through \emph{token indexing and selection}, thanks to the image tokenization process in vision Transformers, thereby avoiding the complex pixel gathering required in convolution layers (\figurename~\ref{fig:arch} (b)).
From an algorithmic design perspective, the recent AdaViT framework \cite{meng2022adavit} introduces a method for adaptively skipping tokens, heads/channels in multi-head attention, and layers, thus enabling dynamic computation across spatial, width, and depth dimensions simultaneously. However, despite theoretical comparisons presented in \cite{meng2022adavit}, the practical efficacy of these paradigms on hardware remains uncertain. This paper leverages the architectural design principles of AdaViT to circumvent the need for foundational redesigns and utilizes our proposed latency predictor to conduct a thorough examination of the practical performance of these dynamic paradigms in vision Transformers.
} 


\begin{figure}
    \centering
    \includegraphics[width=\linewidth]{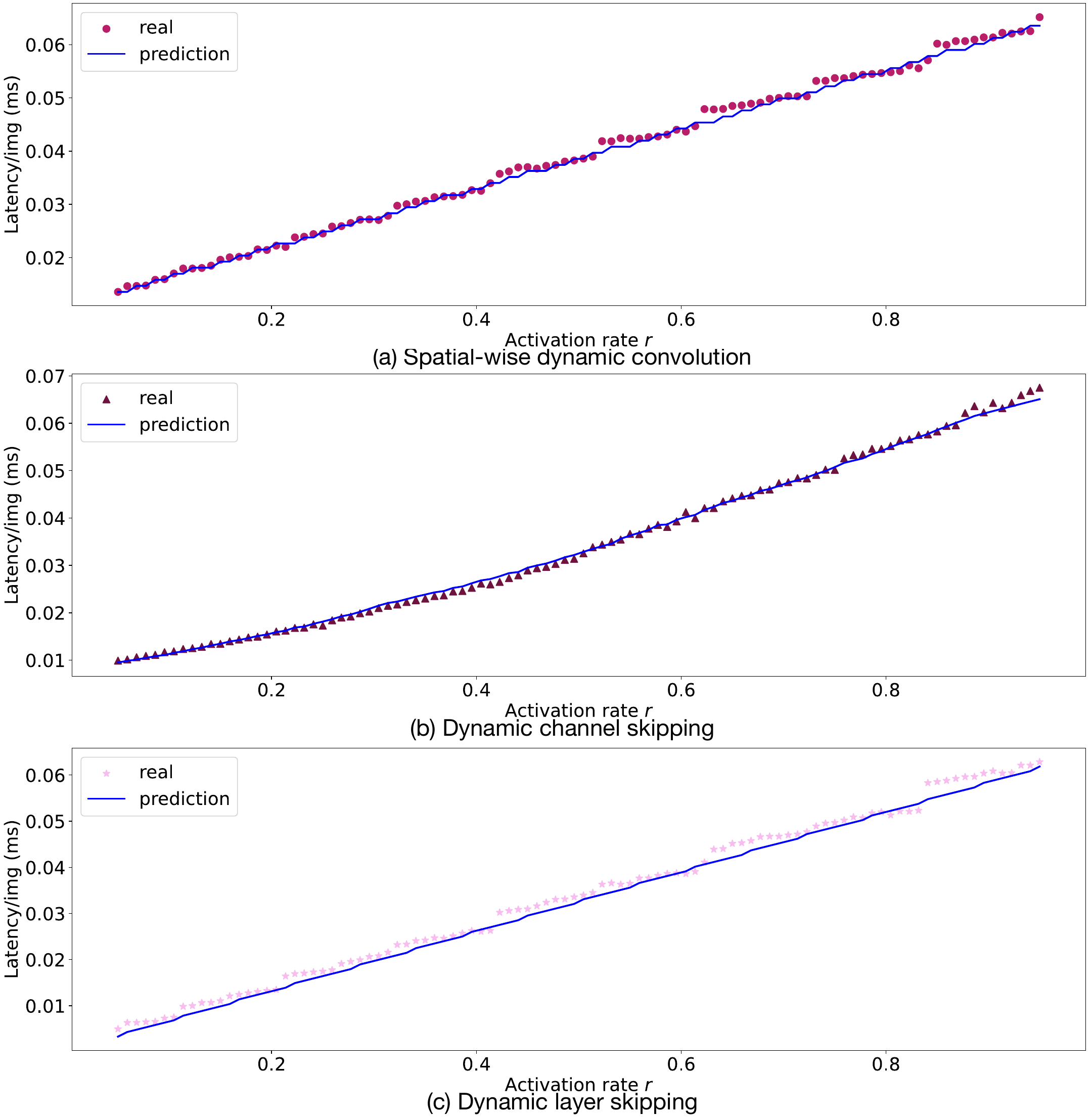} 
    \vskip -0.1in
    \caption{Comparison between the real and predicted latency of a dynamic block in LAUD-ResNet-101.} 
    \label{real_predicted_latency} 
    \vskip -0.25in
\end{figure}
\subsection{Latency predictor}\label{sec_predictor}
As stated before, it is laborious to evaluate the latency of dynamic operators on different hardware platforms. To efficiently seek preferable dynamic paradigms and granularity settings on any target device, we propose a latency prediction model $\mathcal{G}$. Given hardware properties $\mathbf{H}$, layer parameters $\mathbf{P}$, dynamic paradigm $\mathbf{D}$, spatial/channel granularity $S/C$, and activation rates $r^\mathrm{s}/r^\mathrm{c}$, $\mathcal{G}$ directly \emph{predicts} block execution latency $\ell= \mathcal{G}(\mathbf{H}, \mathbf{P}, \mathbf{D}, S, C, r^\mathrm{s}, r^\mathrm{c})$.


\noindent\textbf{Hardware modeling.} We model a device with multiple processing engines (PEs) for parallel computation (\figurename~\ref{hardware_model}). The memory system has three levels \cite{hennessy2011computer}: 1) off-chip memory, 2) on-chip global memory, and 3) memory in PE. In practice, the latency mainly comes from two processes: \emph{data movement} and \emph{parallel computation}: 
\begin{equation}
\ell=\ell_{\mathrm{data}}+\ell_{\mathrm{computation}} + \ell_\mathrm{Const}, 
\end{equation}
where $\ell_\mathrm{Const}$ is a hardware-specific constant. This model accurately predicts both $\ell_{\mathrm{data}}$ and $\ell_{\mathrm{computation}}$, enabling more practical efficiency measurement than FLOPs.

\noindent\textbf{Latency prediction.} Given hardware properties and model parameters, adopting a proper \emph{scheduling strategy} is key to maximizing resource utilization through increased parallelism and reduced memory access. We use Nvidia Cutlass \cite{cutlass} to \emph{search} for the optimal scheduling (tiling and in-PE parallelism configurations) of dynamic operations. The data movement latency can then be easily obtained from data shapes and target device bandwidth. Furthermore, the computation latency is derived from hardware properties. Please refer to Appendix \ref{detailed_latency_predict} for more details.

\noindent\textbf{Empirical validation.} We evaluate the performance of our latency predictor with a ResNet-101 block on an RTX3090 GPU, varying the activation rate $r$. The blue curves represent the predictions, and the scattered dots are obtained via \emph{searching} for a proper scheduling strategy (implemented with custom CUDA code) using Nvidia Cutlass \cite{cutlass}. All the three dynamic paradigms are tested. \figurename~\ref{real_predicted_latency} compares predictions to real GPU testing latency, showing accurate estimates across a wide range of activation rates.



\subsection{Scheduling optimization}\label{sec_schedule_optim}
We use general optimization methods like fusing activation functions and batch normalization (BN) layers into convolution layers.
We also optimize our dynamic convolution blocks as follows.

\noindent\textbf{Operator fusion for spatial maskers}.
As mentioned in Sec.~\ref{sec:arch}, spatial maskers have negligible computation but take the full feature map as input, making them \emph{memory-bounded} (latency is dominated by memory access).
Since the masker shares its input with the first $1\times1$ conv (Masker-Conv1$\times$1 in Figure \ref{fig:arch} (b)), fusing them avoids repeated input reads. 
However, this makes the convolution spatially static, potentially increasing computation. For simplicity, we adopt such operator fusion in all tested models. In practice, we find that operator fusion improves efficiency in most scenarios.

\noindent\textbf{Fusing gather and dynamic convolution}.
Traditional approaches first gather the input pixels of the first dynamic convolution in a block. The gather operation is also a \emph{memory-bounded} operation. Furthermore, when the kernel size exceeds 1$\times$1, input patches overlap, leading to repeated loads/stores.
We fuse gathering into dynamic convolution to reduce the memory access (Gather-Conv3x3 in \figurename~\ref{fig:arch} (b)).

Note that for dynamic channel skipping (\figurename~\ref{fig:arch} (c)), gathering is conducted on convolution kernels rather than features. The weight gather operations is also fused with convolution by our scheduling optimization.

\noindent\textbf{Fusing scatter and add.} Conventional methods scatter the final convolution outputs before the element-wise addition.
We fuse these two operators (Scatter-Add in \figurename~\ref{fig:arch} (b)) to reduce memory access costs. The ablation study in Sec.~\ref{sec_results_latency_pred} validates the effectiveness of the proposed fusing methods.

\noindent\textbf{Batching inference} is enabled by recording patch, location, and sample correspondences during gathering and scattering (\figurename~\ref{fig:arch} (b, c)). Inference with a larger batch size facilitates parallel computation, making latency more dependent on computation versus kernel launching or memory access. See Appendix~\ref{supp_results_latency_pred} for empirical analysis.

\subsection{Training}\label{sec:train}
\noindent\textbf{Optimization of non-differentiable maskers.} The masker modules produce binary variables for discrete decisions, and cannot be directly optimized with back-propagation. 
Following \cite{xie2020spatially,verelst_dynamic_2020,SAR_TIP}, we adopt straight-through Gumbel Softmax \cite{jang2016categorical,maddison2016concrete} for end-to-end training. 
Take spatial-wise dynamic inference as an example, let $\tilde{\mathbf{M}}^\mathrm{s}\!\in\!\mathbb{R}^{H\times W\times 2}$ denote the output of the spatial mask generator $\mathcal{M}^\mathrm{s}$. The decisions are obtained with the argmax function during inference. Training uses a differentiable Softmax approximation:
\begin{equation}\footnotesize
    \hat{\mathbf{M}}^\mathrm{s}=\frac{\exp\left\{\left(\log\left(\mathbf{\tilde{M}}^\mathrm{s}_{:,:,0}\right)+\mathbf{G}_{:,:,0}\right)/\tau\right\}}{\sum_{k=0}^1 \exp\left\{\left(\log\left(\mathbf{\tilde{M}}^\mathrm{s}_{:,:,k}\right)+\mathbf{G}_{:,:,k}\right)/\tau\right\}}\in[0,1]^{H\times W},
\end{equation}
where $\tau$ is the Softmax temperature. Similarly, a channel masker $\mathcal{M}^\mathrm{c}$ produces a $2C$-dimensional vector $\tilde{\mathbf{M}}^\mathrm{c}\in\mathbb{R}^{2C}$, where $C$ is the channel number of the $3\times 3$ convolution in a block. We first reshape $\tilde{\mathbf{M}}^\mathrm{c}$ into the size of $C\times 2$, and apply Gumbel Softmax along the second dimension to produce $\hat{\mathbf{M}}^\mathrm{c}\in[0,1]^C$. Following \cite{verelst_dynamic_2020,SAR_TIP}, we let $\tau$ decay exponentially from 5.0 to 0.1 in training to facilitate the optimization of maskers. 

\noindent\textbf{Training objective.}  
As analyzed in Sec.~\ref{sec:arch}, the FLOPs of each dynamic convolution block can be calculated based on our defined activation rate $r^\mathrm{s}$ (or $r^\mathrm{c}$). Let $F_{\mathrm{dyn}}$ and $F_{\mathrm{stat}}$ denote the overall dynamic and static network FLOPs. We optimize their ratio to approximate a target $0\!<\!t\!<\!1$: $L_{\mathrm{FLOPs}}=(\frac{F_{\mathrm{dyn}}}{F_{\mathrm{stat}}}-t)^2.$
In addition, we define $L_{\mathrm{bounds}}$ as in \cite{verelst_dynamic_2020} to constrain the upper/lower bounds in early training epochs.

We further propose to leverage the static counterparts of our dynamic networks as ``teachers'' to guide the optimization procedure. Let $\mathbf{y}$ and $\mathbf{y}'$ denote the output logits of a dynamic ``student'' model and its static ``teacher'', respectively. Our final loss can be written as
 \begin{equation}\label{eq_loss}\footnotesize
    L = L_{\mathrm{task}} + \alpha (L_\mathrm{FLOPs}+L_{\mathrm{bounds}}) + \beta T^2\cdot \mathrm{KL}(\sigma(\mathbf{y}/T)||\sigma(\mathbf{y}'/T)),
\end{equation}
where $L_\mathrm{task}$ represents the task-related loss, \emph{e.g.}, cross-entropy loss in classification. $\mathrm{KL}(\cdot||\cdot)$ denotes the Kullback–Leibler divergence, and $\alpha,\beta$ are the coefficients balancing these items. We use $\sigma$ to denote the log-Softmax function, and $T$ is the temperature for computing KL-divergence.

\begin{figure*} 
\vskip -0.1in
    \begin{subfigure}[b]{\linewidth}
        \centering
        \includegraphics[width=\linewidth]{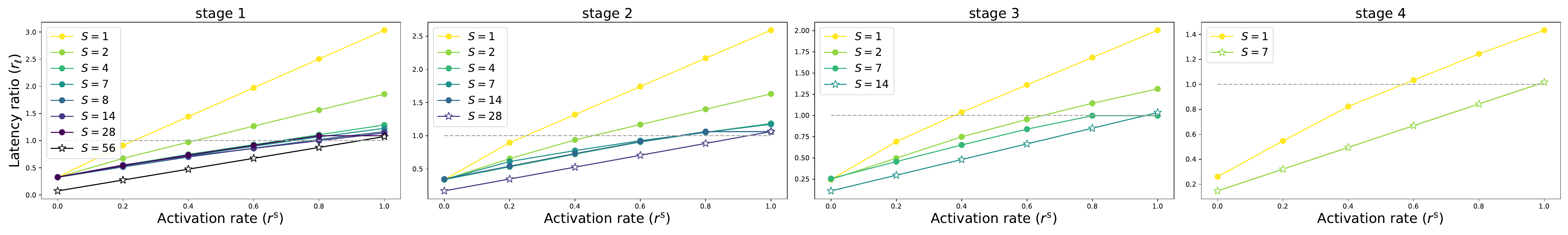} 
        \vskip -0.1in
        \caption{Relationship between the latency ratio $r_{\ell}$ and the activation rate $r^\mathrm{s}$ for $\mathrm{LAUD}^{\mathrm{s/l}}$-ResNet.} 
        \label{fig:r_l_vs_r_s_res_v100}
    \end{subfigure}\hfill
    \begin{subfigure}[b]{\linewidth}
        \centering
        \includegraphics[width=\linewidth]{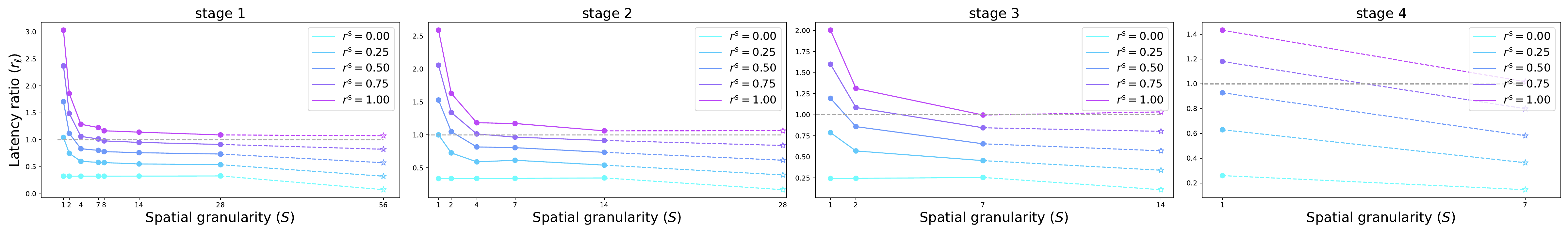} 
        \vskip -0.1in
        \caption{Relationship between the latency ratio $r_{\ell}$ and the spatial granularity $S$ for $\mathrm{LAUD}^{\mathrm{s/l}}$-ResNet.} 
        \label{fig:r_l_vs_S_res_v100}
    \end{subfigure}\hfill
    \begin{subfigure}[b]{\linewidth}
        \centering
        \includegraphics[width=\linewidth]{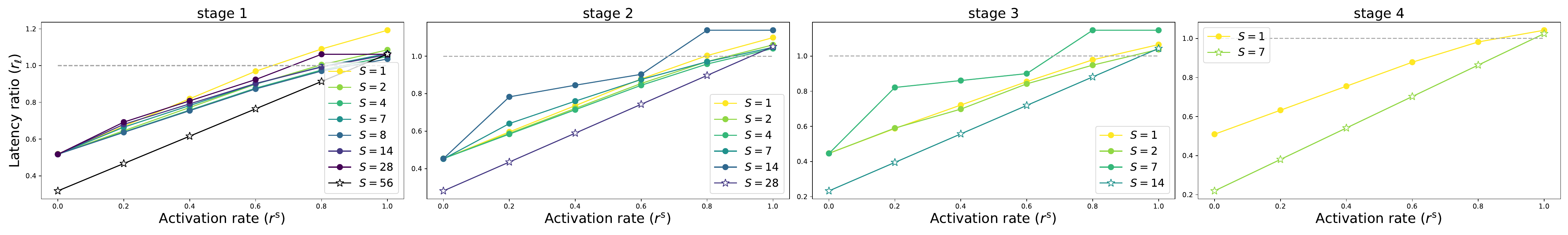} 
        \vskip -0.1in
        \caption{Relationship between the latency ratio $r_{\ell}$ and the activation rate $r^\mathrm{s}$ for $\mathrm{LAUD}^{\mathrm{s/l}}$-RegNeY-800M.} 
        \label{fig:r_l_vs_r_s_reg_tx2}
    \end{subfigure}\hfill
    \begin{subfigure}[b]{\linewidth}
        \centering
        \includegraphics[width=\linewidth]{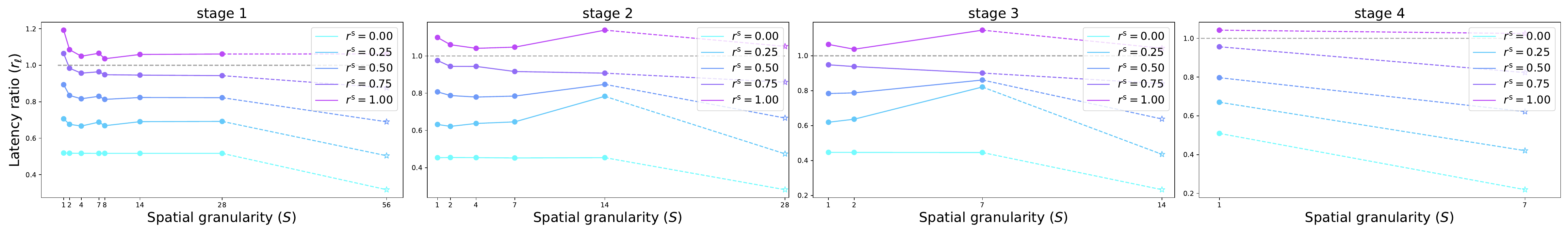} 
        \vskip -0.1in
        \caption{Relationship between the latency ratio $r_{\ell}$ and the spatial granularity $S$ for $\mathrm{LAUD}^{\mathrm{s/l}}$-RegNeY-800M.} 
        \label{fig:r_l_vs_S_reg_tx2}
    \end{subfigure}
\vskip -0.1in
\caption{Latency prediction results for $\mathrm{LAUD}^{\mathrm{s/l}}$-ResNet blocks on the Nvidia Tesla V100 GPU (a, b) and $\mathrm{LAUD}^{\mathrm{s/l}}$-RegNetY-800MF blocks on the Nvidia Jetson TX2 GPU (c, d). The circle markers ($\bullet$) represent spatial-wise dynamic computation, and the star markers ($\star$) denote layer skipping, which is implemented via the largest granularity $S$ in each stage.}
\label{fig:r_l_vs_r_s_and_S}
\vskip -0.2in
\end{figure*}

\section{Experiments}
In this section, we first introduce the experiment settings in Sec.~\ref{sec_setup}. Then the latency of different granularity settings are analyzed in Sec.~\ref{sec_results_latency_pred}. The performance of our \nameShort~on ImageNet is further evaluated in Sec.~\ref{sec_IN_results}, followed by the visualization results in Sec.~\ref{sec_vis}. We finally validate our method on the object detection and instance segmentation tasks (Sec.~\ref{sec_det}). For simplicity, we add ``$\mathrm{LAUD}^{\mathrm{s/c/l}}$-'' as a prefix before model names to denote our \nameShort~with different dynamic paradigms (s for spatial, c for channel and l for layer), \emph{e.g.}, $\mathrm{LAUD}^{\mathrm{s}}$-ResNet-50. 

\subsection{Experiment setup}\label{sec_setup}
\noindent\textbf{Image classification} experiments are conducted on the ImageNet \cite{deng2009imagenet} dataset.  
\textcolor{black}{We implement our LAUDNet on five representative architectures extending up to a broad spectrum of computational costs: four CNNs (ResNet-50, ResNet-101 \cite{he2016resnet}, RegNetY-400M, RegNetY-800M \cite{radosavovic2020designing}) and a vision Transformer, T2T-ViT~\cite{yuan2021tokens}. Different training settings are used for CNNs and Transformers. For CNNs,} As per the established methodology in \cite{verelst_dynamic_2020}, we initialize the backbone parameter from a torchvision pre-trained checkpoint (\url{https://pytorch.org/vision/stable/models.html}), and finetune the whole network for 100 epochs employing the loss function in Eq.~(\ref{eq_loss}). We fix $\alpha\!=\!10,\beta\!=\!0.5$ and $T\!=\!4.0$ for all dynamic models. Note that we adopt the pretrain-fintune paradigm mainly to reduce the training cost, as Gumbel Softmax usually requires longer training for convergence. \textcolor{black}{For our study on T2T-ViT, we use the same setup as described in AdaViT~\cite{meng2022adavit} and evaluate the efficiency of its various dynamic inference methods through our latency predictor.}

\noindent\textbf{Latency prediction.} We evaluate our LAUDNet on various types of hardware platforms, including two server GPUs (Tesla V100 and RTX3090), a desktop GPU (RTX3060) and two edge devices (\emph{e.g.}, Jetson TX2 and Nvidia Nano). The major properties considered by our latency prediction model include the number of processing engines (\#PE), the floating-point computation in a processing engine (\#FP32), the frequency and the bandwidth. It can be observed from Tab.~\ref{tab_hardware_property} that server GPUs generally have a larger \#PE than IoT devices. If not stated otherwise, the batch size is set as 128 for V100, RTX3090 and RTX3060 GPUs. On edge devices TX2 and Nano, tesing batch size is fixed as 1.

More details are provided in Appendix~\ref{sec_detailed_settings}.

\begin{figure*} 
 \begin{subfigure}[b]{\linewidth}
    \centering
    \includegraphics[width=\linewidth]{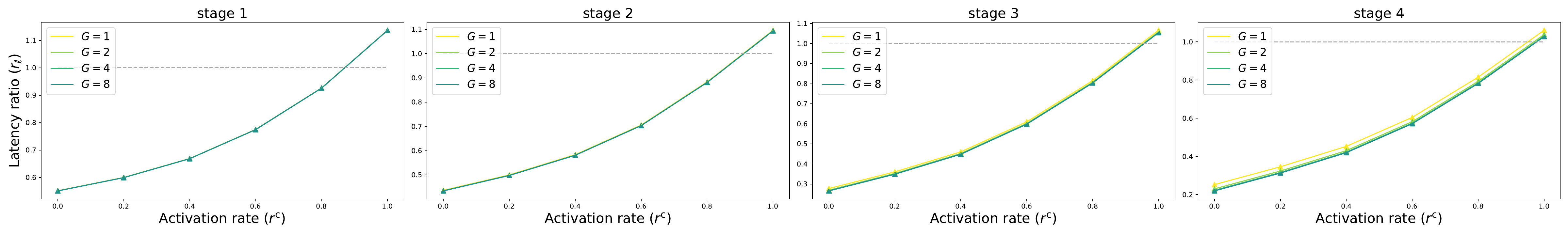} 
    \vskip -0.1in
    \caption{Relationship between the latency ratio $r_{\ell}$ and the activation rate $r^\mathrm{c}$ for $\mathrm{LAUD}^{\mathrm{c}}$-ResNet.} 
    \label{fig:r_l_vs_r_c_res_v100}
\end{subfigure}\hfill
\begin{subfigure}[b]{\linewidth}
    \centering
    \includegraphics[width=\linewidth]{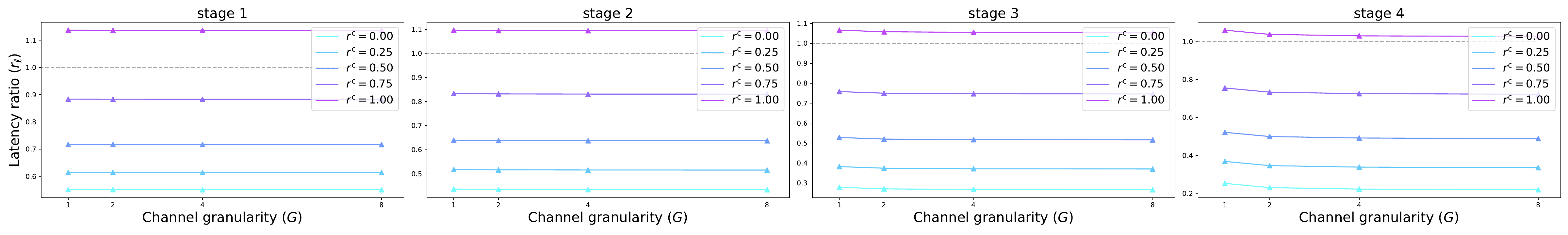} 
    \vskip -0.1in
    \caption{Relationship between the latency ratio $r_{\ell}$ and the channel granularity $G$ for $\mathrm{LAUD}^{\mathrm{c}}$-ResNet.} 
    \label{fig:r_l_vs_G_res_v100}
\end{subfigure}
\vskip -0.1in
\caption{Latency prediction results for $\mathrm{LAUD}^{\mathrm{c}}$-ResNet blocks on the Nvidia Tesla V100 GPU.}
\label{fig:r_l_vs_r_c_and_G}
\vskip -0.2in
\end{figure*}

\subsection{Latency prediction results}\label{sec_results_latency_pred}
This subsection presents the latency prediction results of dynamic convolutional blocks using two distinct backbones: ResNet-50 \cite{he2016resnet} (on V100) and RegNetY-800MF \cite{radosavovic2020designing} (on TX2). Each block features a bottleneck structure with varying channel numbers and convolution groups, and the RegNetY employs Squeeze-and-Excitation (SE) \cite{hu2018squeeze} modules. We define $\ell_{\mathrm{dyn}}$ as the latency of a dynamic convolutional block and $\ell_{\mathrm{stat}}$ as the latency of a static block. The ratio of the two is denoted as $r_{\ell}=\frac{\ell_{\mathrm{dyn}}}{\ell_{\mathrm{stat}}}$, with a realistic speedup being achieved when $r_{\ell}<1$. 

\noindent\textbf{Effect of spatial granularity.} The primary objective here is to investigate how the \emph{granularity} of dynamic computation impacts the latency ratio $r_{\ell}$. We explore the correlation between $r_{\ell}$ and the activation rate $r^\mathrm{s}$ (refer to Sec.~\ref{sec:arch}) for varying \emph{granularity} settings. The results in \figurename~\ref{fig:r_l_vs_r_s_res_v100} (ResNet on V100) and \figurename~\ref{fig:r_l_vs_r_s_reg_tx2} (RegNetY-800M on TX2) demonstrate that:
\begin{itemize}
    \item Despite the implementation of our optimized scheduling strategies, pixel-level dynamic convolution ($S$=1) does not consistently enhance practical efficiency. This approach to fine-grained adaptive inference has been adopted in previous works \cite{verelst_dynamic_2020,xie2020spatially,colleman2021processor}. Our findings help elucidate why these studies only managed to achieve realistic speedup on less potent CPUs \cite{xie2020spatially} or specialized devices \cite{colleman2021processor};
    \item By contrast, a coarse granularity setting ($S>1$) significantly mitigates this issue across both devices. Realistic speedup ($r_{\ell}<1$) is attainable with larger activation values ($r^\mathrm{s}$) when $S>1$.
\end{itemize}

\begin{table}
  \caption{Ablation studies on operator fusion.}
  \vskip -0.1in
  \label{ablation_op_fusion}
  \centering
  \begin{tabular}{ccccc}
    \toprule
    \multirow{2}{*}{Masker-Conv} & \multirow{2}{*}{Gather-Conv} & \multirow{2}{*}{Scatter-Add}     & \multicolumn{2}{c}{Latency (\textmu s)} \\
    & & & V100 & \textcolor{black}{TX2} \\
    \midrule
    \xmark & \xmark & \xmark & 162.4 & \textcolor{black}{1084.5} \\
    \cmark & \xmark & \xmark & 135.1 & \textcolor{black}{1072.3} \\
    \cmark & \cmark & \xmark & 131.7 & \textcolor{black}{1024.7} \\
    \cellcolor{lightgray!40}\cmark & \cellcolor{lightgray!40}\cmark & \cellcolor{lightgray!40}\cmark & \cellcolor{lightgray!40}\textbf{118.3} & \cellcolor{lightgray!40}\textbf{\textcolor{black}{859.7}} \\
    \bottomrule
  \end{tabular}
  \vskip -0.2in
\end{table}

The latency prediction results are further used to determine preferable spatial granularity settings for the first 3 stages. Note that for the final stage where the feature resolution is $7\!\times\! 7$, $S\!=\!1$ and $S\!=\!7$ correspond to two distinct dynamic paradigms (spatially adaptive inference and layer skipping). The relationship curves between $r_{\ell}$ and $S$ depicted in \figurename~\ref{fig:r_l_vs_S_res_v100} (ResNet on V100) and \figurename~\ref{fig:r_l_vs_S_reg_tx2} (RegNetY-800M on TX2) reveal the following: 
\begin{itemize}
    \item The latency ratio $r_{\ell}$ generally decreases as $S$ increases for a given $r$ on V100;
    \item An excessively large $S$ (indicating less flexible adaptive inference) provides negligible improvement on both devices. In particular, increasing $S$ from 7 to 14 in the second stage of LAUD-RegNetY-800MF on TX2 detrimentally impacts efficiency. This is hypothesized to be due to the oversized patch size causing additional memory access costs on this device, which has fewer processing engines (PEs);
    \item  Layer skipping (marked by $\star$) consistently outperforms spatial-wise dynamic computation (marked by $\bullet$). We will analyze their performance across various vision tasks in Sec.~\ref{sec_IN_results} and Sec.~\ref{sec_det}.
\end{itemize}

Based on these results, we can strike a balance between flexibility and efficiency by choosing suitable $S$ for different models and devices. For instance, we can simply set $S^{\mathrm{net}}$=4-4-2-1\footnote{We use this form to represent the $S$ settings for the 4 network stages.} in a LAUD$^{\mathrm{s}}$-ResNet-50 to achieve realistic speedup.

\noindent\textbf{Effect of channel granularity.} We further investigate how the channel granularity $G$ influences the realistic latency of channel-skipping dynamic models. Using $\mathrm{LAUD}^{\mathrm{c}}$-ResNet as an example, results presented in \figurename~\ref{fig:r_l_vs_r_c_and_G} show that the performance of channel skipping is less sensitive to the channel granularity $G$. Setting $G\!=\!2$ improves efficiency only in deeper stages, while extending $G$ beyond 2 offers diminishing benefits. This aligns with our understanding that channel skipping requires more regular operations compared to spatially sparse convolution, implying that $G\!=\!1$ can already employ impactful speedup. Moreover, the curves in \figurename~\ref{fig:r_l_vs_r_c_res_v100} are generally convex, since the computation of the $3\times 3$ convolution is quadratic in relation to $r^\mathrm{c}$ (Sec.~\ref{sec:arch}).

\noindent\textbf{Ablation study of operator fusion.} \textcolor{black}{In exploring the impact of our operator fusion, as detailed in Sec.~\ref{sec_schedule_optim}, we focus on a convolutional block from the initial stage of LAUD$^\mathrm{s}$-ResNet-50 ($S$=4, $r^\mathrm{s}$=0.6) for our case study. The findings, presented in Tab.~\ref{ablation_op_fusion}, reveal that operator fusion consistently aids in lowering the practical latency across different computing environments by minimizing memory access overhead. Notably, the fusion between the masker and the first convolution emerges as a significant factor in reducing latency on the server-end V100. In contrast, combining scattering and addition operations plays a pivotal role in latency reduction on the edge device TX2.}


\begin{figure} 
    \centering
    \includegraphics[width=\linewidth]{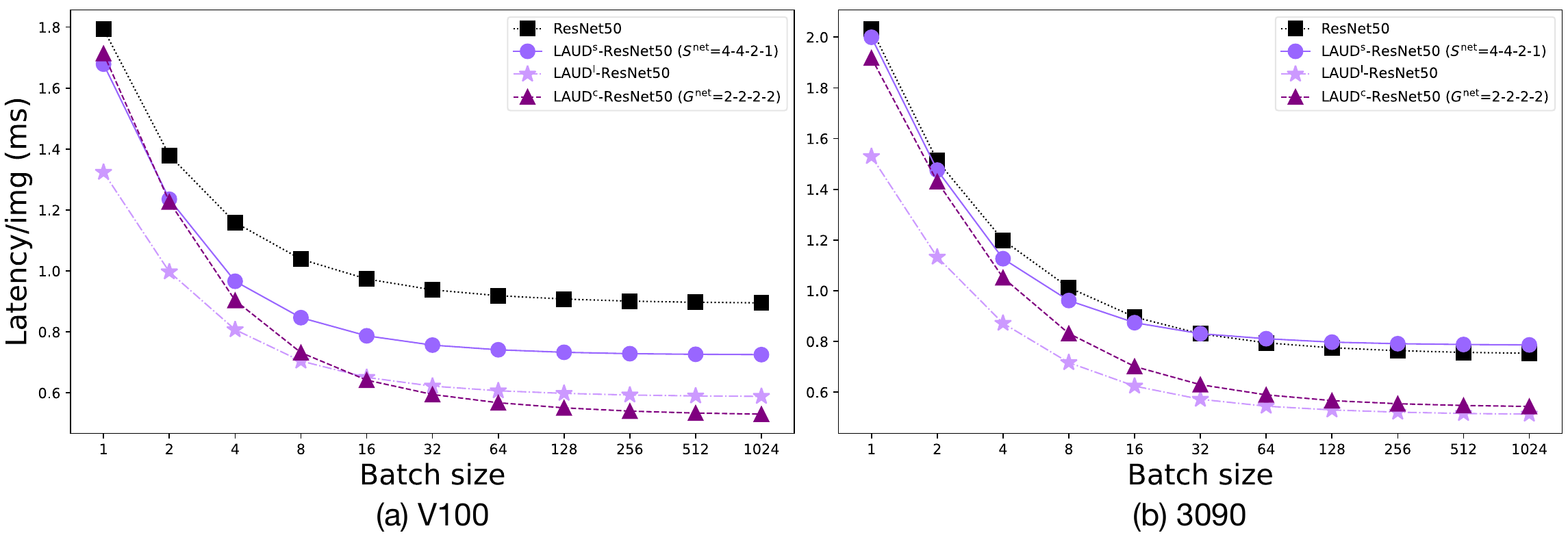} 
    \vskip -0.1in
    \caption{Relationship between the latency per image and batch size of LAUD-ResNet-50 on V100 (a) and 3090 (b) GPUs.} 
    \label{fig:latency_bs}
    \vskip -0.2in
\end{figure}

\noindent\textbf{Ablation study of batch size.} To establish a suitable testing batch size, we graph the relationship between latency per image and batch size for LAUD-ResNet-50 in \figurename~\ref{fig:latency_bs}. Two server-end GPUs (V100 and RTX3090) are tested. The results highlight that latency diminishes with an increase in batch size, eventually reaching a stable plateau when the batch size exceeds 128 on both platforms. This is comprehensible since a larger batch size favors enhanced computation parallelism, resulting in latency becoming more dependent on theoretical computation. The results on the desktop-level GPU, RTX3060 (\figurename~\ref{fig:latency_bs_3060} in Appendix~\ref{supp_results_latency_pred}), show a similar phenomenon. Based on these observations, we report the latency on server-end and desktop-level GPUs with a batch size of 128 henceforth.

\subsection{ImageNet classification}\label{sec_IN_results}

\subsubsection{Comparison of spatial/channel granularities}
\begin{figure} 
    \centering
    \includegraphics[width=\linewidth]{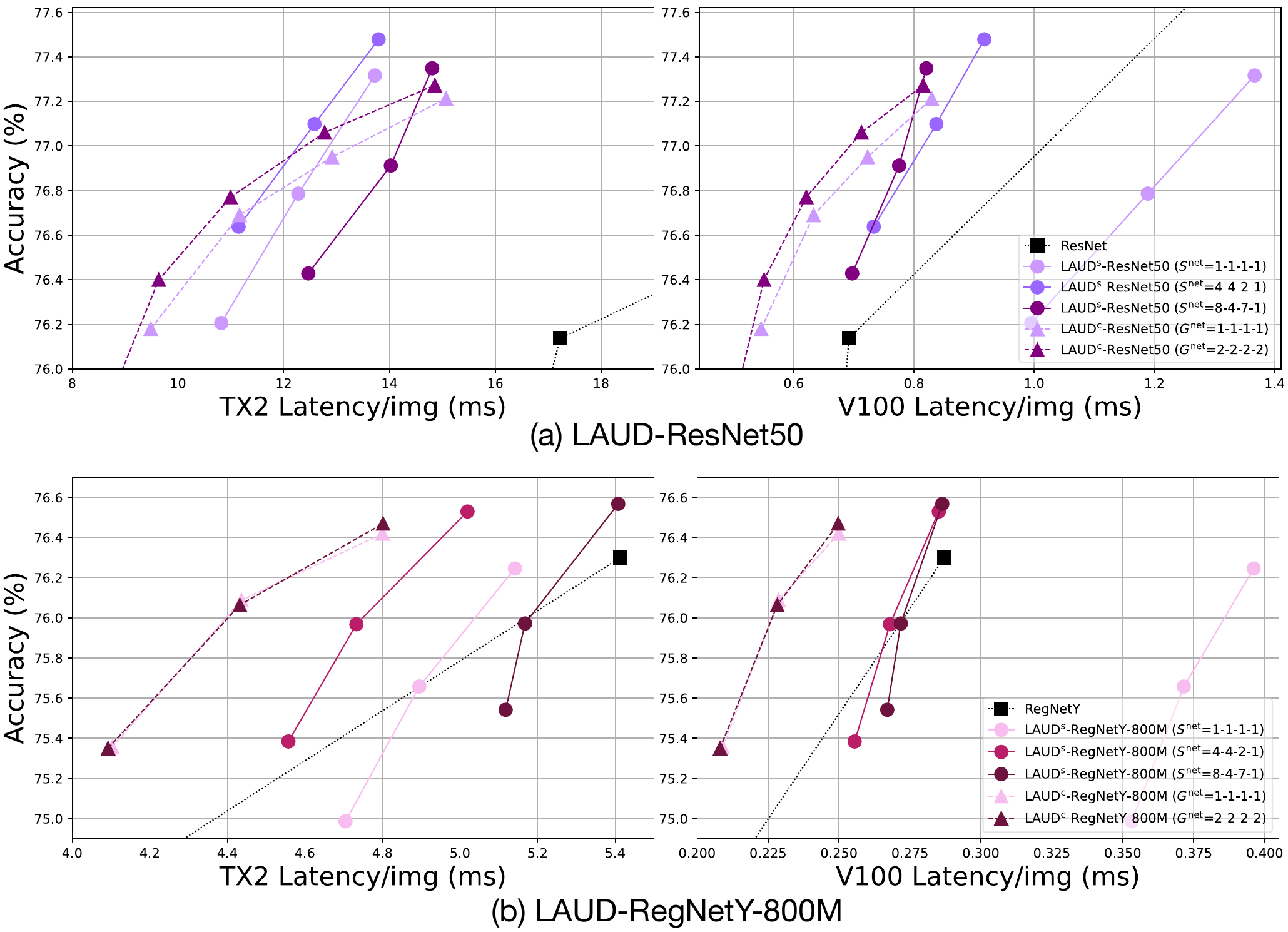} 
    \vskip -0.1in
    \caption{Comparison of different granularities ($S$ and $G$) in LAUD-ResNet-50 (a) and LAUD-RegNetY-800M (b). The latency on TX2 (left) and V100 (right) are presented.} 
    \label{fig:compare_S_G}
    \vskip -0.1in
\end{figure}

We begin by comparing different granularities for spatial and channel-wise dynamic computation. Based on the analysis in Sec.~\ref{sec_results_latency_pred}, the candidates for spatial and channel granularities are $S^{\mathrm{net}}\in$\{1-1-1-1, 4-4-2-1, 8-4-7-1\} and $G^{\mathrm{net}}\in$\{1-1-1-1, 2-2-2-2\} respectively. We select ResNet-50 and RegNetY-800M as backbones, and compare various settings on TX2 and V100. The results in \figurename~\ref{fig:compare_S_G} reveal that:
\begin{itemize}
    \item Regarding spatially dynamic computation, the optimal granularity $S^{\mathrm{net}}$ is contingent on both network structures and hardware devices. For instance, $S^{\mathrm{net}}$=8-4-7-1 achieves a preferable performance on V100 for both models, yet incurs substantial inefficiency on TX2. This corresponds to our results in \figurename~\ref{fig:r_l_vs_r_s_and_S}. 
    \item Elevating the channel granularity $G$ from 1 to 2 does yield sort of speedup for ResNet-50, but renders comparable performance in the case of RegNetY-800M. We hypothesize that a larger $G$ is only beneficial for models with more extensive channel numbers, which also aligns with observations from \figurename~\ref{fig:r_l_vs_r_c_and_G}.
\end{itemize}

\subsubsection{Comparison of dynamic paradigms}\label{sec:main_results}
\begin{figure*} 
    \centering
    \includegraphics[width=\linewidth]{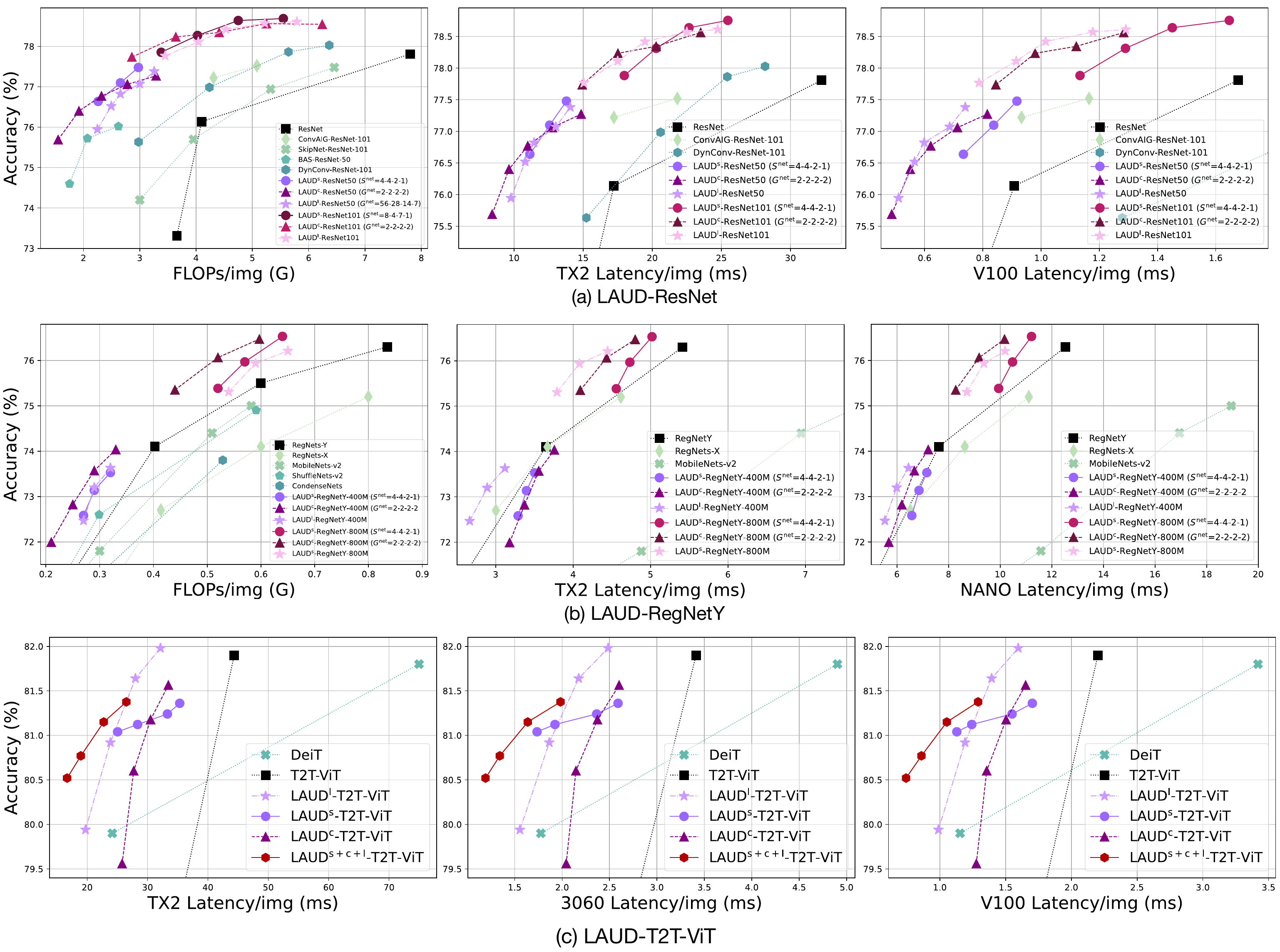} 
    \vskip -0.1in
    \caption{\textcolor{black}{Main results of LAUDNet implemented on ResNet (a), RegNetY (b) and T2T-ViT (c)}.} 
    \label{fig:main_results}
    \vskip -0.1in
\end{figure*}

Having decided on the optimal granularities, we submit different dynamic paradigms to a more detailed comparison. Additionally, our LAUDNet is compared to various competitive baselines. The findings are illustrated in \figurename~\ref{fig:main_results}.

\begin{figure*}
\centering
\includegraphics[width=\linewidth]{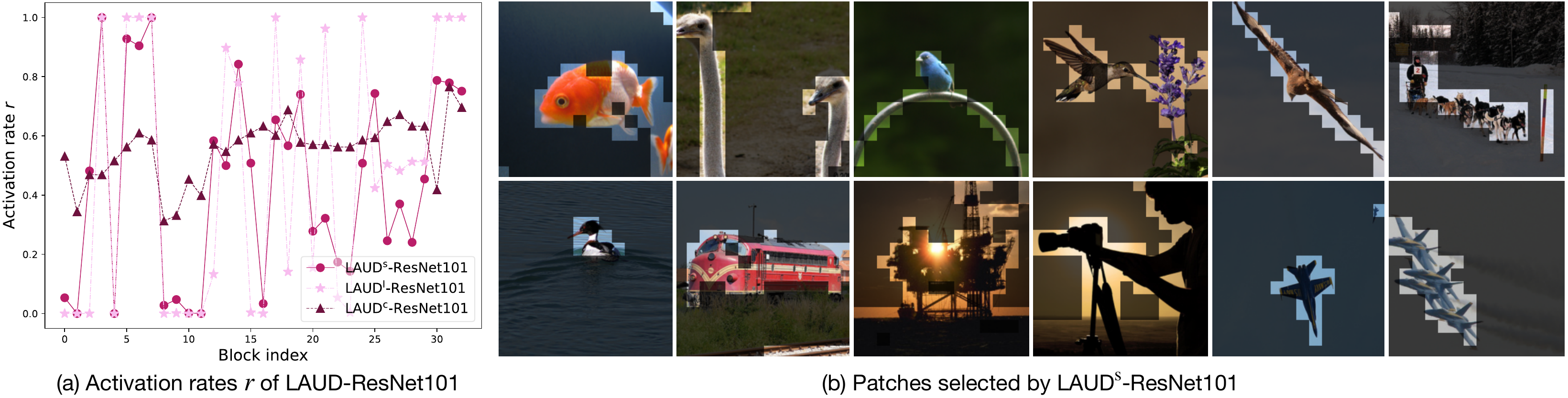}
\vskip -0.1in
\caption{Visualization results of activation rates $r^\mathrm{s/c/l}$ and selected patches by LAUD$^\mathrm{s}$-ResNet-101.}
\label{fig_vis}
\vskip -0.2in
\end{figure*}

\noindent\textbf{Standard baseline comparison: ResNets.} The compared baselines include various types of dynamic inference approaches: 1) layer skipping (SkipNet \cite{wang2018skipnet} and Conv-AIG \cite{veit2018convolutional}); 2) channel skipping (BAS \cite{bejnordi2019batch}); and 3) pixel-level spatial-wise dynamic network (DynConv \cite{verelst_dynamic_2020}). For our \nameShort, we select the best granularity settings for spatial-wise and channel-wise dynamic inference. Layer skipping implemented in our framework is also included. We set training targets (cf. Sec.~\ref{sec:train}) $t\!\in\!\{0,4,\cdots,0.8\}$ for our dynamic models to evaluate their performance across different sparsity regimes. We apply scheduling optimization (Sec.~\ref{sec_schedule_optim}) uniformly across all models \cite{veit2018convolutional,verelst_dynamic_2020} for a fair comparison.

\begin{table*}
  \caption{Object detection results on the COCO dataset.}
  \vskip -0.2in
  \label{tab:coco}
  \begin{center}
  \begin{tabular}{c c c c c c c c c}
  \toprule
  Detection & \multirow{2}{*}{Backbone} & Backbone & \multicolumn{5}{c}{Backbone Latency (ms)} & \multirow{2}{*}{mAP (\%)} \\
  \cmidrule{4-8}
  Framework & & FLOPs (G) & V100 & 3090 & 3060 & TX2 & Nano &\\
  \midrule
   \multirow{10}{*}{Faster R-CNN} & ResNet-101 (Baseline)  & 141.2 & 33.9 & 29.8 & 44.8 & 586.4 & 1600.4 & 39.4 \\ 
   \cmidrule{2-9}
   & \cellcolor{lightgray!40}LAUD$^\mathrm{s}$-ResNet-101 ($S^{\mathrm{net}}$=4-4-2-1, $t$=0.6) & \cellcolor{lightgray!40}90.7  & \cellcolor{lightgray!40}32.4 & 
   \cellcolor{lightgray!40}36.8 & 
   \cellcolor{lightgray!40}40.4 &
   \cellcolor{lightgray!40}402.2 &
   \cellcolor{lightgray!40}1082.4  &
   \cellcolor{lightgray!40}\textbf{40.3} \\
   & \cellcolor{lightgray!40}LAUD$^\mathrm{s}$-ResNet-101 ($S^{\mathrm{net}}$=4-4-7-1, $t$=0.5) & \cellcolor{lightgray!40}79.5  & \cellcolor{lightgray!40}30.4	& 
   \cellcolor{lightgray!40}29.4	& 
   \cellcolor{lightgray!40}38.2	& 
   \cellcolor{lightgray!40}390.4    & 
   \cellcolor{lightgray!40}1050.7 & 
   \cellcolor{lightgray!40}40.0 \\
   & \cellcolor{lightgray!40}LAUD$^\mathrm{s}$-ResNet-101 ($S^{\mathrm{net}}$=4-4-7-1, $t$=0.4) & \cellcolor{lightgray!40}\textbf{67.9} &
   \cellcolor{lightgray!40}27.4	& 
   \cellcolor{lightgray!40}26.2	& 
   \cellcolor{lightgray!40}34.5	& 
   \cellcolor{lightgray!40}340.0& 
   \cellcolor{lightgray!40}911.4 & 
   \cellcolor{lightgray!40}39.5 \\
   \cmidrule{2-9}

    & \cellcolor{lightgray!20}LAUD$^\mathrm{c}$-ResNet-101 ($C^{\mathrm{net}}$=2-2-2-2, $t$=0.8)	& \cellcolor{lightgray!20}112.37	& 
    \cellcolor{lightgray!20}30.6	& 
    \cellcolor{lightgray!20}30.0		& 
    \cellcolor{lightgray!20}42.0	  & 
    \cellcolor{lightgray!20}471.6	  & 
    \cellcolor{lightgray!20}1264.3	  & 
    \cellcolor{lightgray!20}40.2 \\
    & \cellcolor{lightgray!20}LAUD$^\mathrm{c}$-ResNet-101 ($C^{\mathrm{net}}$=2-2-2-2, $t$=0.7)	& \cellcolor{lightgray!20}96.42	& 
    \cellcolor{lightgray!20}27.9	& 
    \cellcolor{lightgray!20}27.3		& 
    \cellcolor{lightgray!20}37.8		& 
    \cellcolor{lightgray!20}400.4	     & 
    \cellcolor{lightgray!20}1065.4	  & 
    \cellcolor{lightgray!20}40.0 \\
    & \cellcolor{lightgray!20}LAUD$^\mathrm{c}$-ResNet-101 ($C^{\mathrm{net}}$=2-2-2-2, $t$=0.6)	& \cellcolor{lightgray!20}80.73	& 
    \cellcolor{lightgray!20}23.9	& 
    \cellcolor{lightgray!20}24.6		& 
    \cellcolor{lightgray!20}33.9		& 
    \cellcolor{lightgray!20}335.7	     & 
    \cellcolor{lightgray!20}\textbf{884.0}	  & 
    \cellcolor{lightgray!20}39.7 \\
   \cmidrule{2-9}

    & \cellcolor{lightgray!60}LAUD$^\mathrm{l}$-ResNet-101 ($t$=0.5)  &\cellcolor{lightgray!60}97.97	&\cellcolor{lightgray!60}24.2	&
    \cellcolor{lightgray!60}22.1		&
    \cellcolor{lightgray!60}32.2		&
    \cellcolor{lightgray!60}409.2	   &
    \cellcolor{lightgray!60}1114.1 &
    \cellcolor{lightgray!60}40.2 \\
    & \cellcolor{lightgray!60}LAUD$^\mathrm{l}$-ResNet-101 ($t$=0.4)  &\cellcolor{lightgray!60}86.71 &
    \cellcolor{lightgray!60}\textbf{19.8}	&
    \cellcolor{lightgray!60}\textbf{18.2}		&
    \cellcolor{lightgray!60}\textbf{26.5}		&
    \cellcolor{lightgray!60}\textbf{331.2}	   &
    \cellcolor{lightgray!60}899.9 &
    \cellcolor{lightgray!60}39.5 \\

   \midrule
   \multirow{7}{*}{RetinaNet} & ResNet-101 (Baseline)     & 141.2 &  33.9 & 29.8 & 44.8 & 586.4 & 1600.4 & 38.5 \\ 
   \cmidrule{2-9}
   & \cellcolor{lightgray!40}LAUD$^\mathrm{s}$-ResNet-101 ($S^{\mathrm{net}}$=4-4-2-1, $t$=0.5) & \cellcolor{lightgray!40}77.8   &
   \cellcolor{lightgray!40}29.0	&
   \cellcolor{lightgray!40}32.7	  &
   \cellcolor{lightgray!40}36.7	  &
   \cellcolor{lightgray!40}350.1          & 
   \cellcolor{lightgray!40}937.1          & 
   \cellcolor{lightgray!40}39.3 \\
   & \cellcolor{lightgray!40}LAUD$^\mathrm{s}$-ResNet-101 ($S^{\mathrm{net}}$=4-4-7-1, $t$=0.4) & \cellcolor{lightgray!40}\textbf{66.4}  &
   \cellcolor{lightgray!40}28.1	 &
   \cellcolor{lightgray!40}26.0	 &
   \cellcolor{lightgray!40}35.2	   &
   \cellcolor{lightgray!40}335.0   &
   \cellcolor{lightgray!40}897.1   &
   \cellcolor{lightgray!40}38.9     \\
   \cmidrule{2-9}
   & \cellcolor{lightgray!20}LAUD$^\mathrm{c}$-ResNet-101 ($C^{\mathrm{net}}$=2-2-2-2, $t$=0.6) 
   &\cellcolor{lightgray!20}79.6    &
   \cellcolor{lightgray!20}23.7	 &
   \cellcolor{lightgray!20}24.4	     &
   \cellcolor{lightgray!20}33.7	     &
   \cellcolor{lightgray!20}331.2   &
   \cellcolor{lightgray!20}871.4   &
   \cellcolor{lightgray!20}39.3     \\
   & \cellcolor{lightgray!20}LAUD$^\mathrm{c}$-ResNet-101 ($C^{\mathrm{net}}$=2-2-2-2, $t$=0.5) & \cellcolor{lightgray!20}65.5    &
   \cellcolor{lightgray!20}20.9	 &
   \cellcolor{lightgray!20}22.1	    &
   \cellcolor{lightgray!20}30.4	   &
   \cellcolor{lightgray!20}\textbf{278.7}   &
   \cellcolor{lightgray!20}\textbf{724.6}	   &
   \cellcolor{lightgray!20}38.5    \\
   \cmidrule{2-9}
   & \cellcolor{lightgray!60}LAUD$^\mathrm{l}$-ResNet-101 ($t$=0.5) 
   &\cellcolor{lightgray!60}95.1   
   &\cellcolor{lightgray!60}23.6	
   &\cellcolor{lightgray!60}21.5	
   &\cellcolor{lightgray!60}31.4	
   &\cellcolor{lightgray!60}397.7        
   &\cellcolor{lightgray!60}1082.5
   &\cellcolor{lightgray!60}\textbf{39.4}     \\
   & \cellcolor{lightgray!60}LAUD$^\mathrm{l}$-ResNet-101 ($t$=0.3) 
   & \cellcolor{lightgray!60}74.4    
   &\cellcolor{lightgray!60}\textbf{18.7}
   &\cellcolor{lightgray!60}\textbf{17.3}	
   &\cellcolor{lightgray!60}\textbf{25.0	}
   &\cellcolor{lightgray!60}311.4  
   &\cellcolor{lightgray!60}846.3
   &\cellcolor{lightgray!60}38.6      \\

   \midrule
   \multirow{1}{*}{\textcolor{black}{Deformable-DETR}} & \multirow{4}{*}{\textcolor{black}{ResNet-50 (Baseline)}}     & \multirow{4}{*}{\textcolor{black}{73.3}} & \multirow{4}{*}{\textcolor{black}{18.4}} & \multirow{4}{*}{\textcolor{black}{16.2}}  & \multirow{4}{*}{\textcolor{black}{25.0}} & \multirow{4}{*}{\textcolor{black}{313.6}} & \multirow{4}{*}{\textcolor{black}{851.4}} & \textcolor{black}{46.9} \\ 

   \multirow{1}{*}{\textcolor{black}{DINO-DETR}} &     & &  &   &  &  & & \textcolor{black}{49.0} \\ 

   \multirow{1}{*}{\textcolor{black}{Rank-DETR}}   &     & &  &   &  &  & & \textcolor{black}{50.2} \\ 

   \multirow{1}{*}{\textcolor{black}{Stable-DINO}}      &     & &  &   &  &  & & \textcolor{black}{50.4} \\ 

 \midrule
   
    \multirow{18}{*}{\textcolor{black}{DDQ-DETR}}  &   \textcolor{black}{ResNet-50 (Baseline)} &  \textcolor{black}{73.3} & \textcolor{black}{18.4} & \textcolor{black}{16.2}  & \textcolor{black}{25.0} & \textcolor{black}{313.6} & \textcolor{black}{851.4}    & \textcolor{black}{51.1}  \\ 

    \cmidrule{2-9}

   & \cellcolor{lightgray!40}\textcolor{black}{LAUD$^\mathrm{s}$-ResNet-50 ($S^{\mathrm{net}}$=8-4-7-1, $t$=0.6)} & \cellcolor{lightgray!40}\textcolor{black}{61.9} &
   \cellcolor{lightgray!40}\textcolor{black}{21.3} &
   \cellcolor{lightgray!40}\textcolor{black}{22.7}  &
   \cellcolor{lightgray!40}\textcolor{black}{27.6} &
   \cellcolor{lightgray!40}\textcolor{black}{287.9}       & 
   \cellcolor{lightgray!40}\textcolor{black}{776.4}         & 
   \cellcolor{lightgray!40}\textcolor{black}{\textbf{51.3}} \\

   & \cellcolor{lightgray!40}\textcolor{black}{LAUD$^\mathrm{s}$-ResNet-50 ($S^{\mathrm{net}}$=8-4-7-1, $t$=0.5)} & \cellcolor{lightgray!40}\textcolor{black}{56.8} &
   \cellcolor{lightgray!40}\textcolor{black}{19.9} &
   \cellcolor{lightgray!40}\textcolor{black}{20.9}  &
   \cellcolor{lightgray!40}\textcolor{black}{25.8} &
   \cellcolor{lightgray!40}\textcolor{black}{264.5}       & 
   \cellcolor{lightgray!40}\textcolor{black}{711.6}         & 
   \cellcolor{lightgray!40}\textcolor{black}{51.1} \\

    \cmidrule{2-9}

   & \cellcolor{lightgray!20}\textcolor{black}{LAUD$^\mathrm{c}$-ResNet-50 ($C^{\mathrm{net}}$=2-2-2-2, $t$=0.7)} 
   &\cellcolor{lightgray!20}\textcolor{black}{50.3}   &
   \cellcolor{lightgray!20}\textcolor{black}{15.0} &
   \cellcolor{lightgray!20}\textcolor{black}{15.2}    &
   \cellcolor{lightgray!20}\textcolor{black}{21.7}     &
   \cellcolor{lightgray!20}\textcolor{black}{222.0} &
   \cellcolor{lightgray!20}\textcolor{black}{581.1}  &
   \cellcolor{lightgray!20}\textcolor{black}{50.7}    \\

   & \cellcolor{lightgray!20}\textcolor{black}{LAUD$^\mathrm{c}$-ResNet-50 ($C^{\mathrm{net}}$=2-2-2-2, $t$=0.6)} 
   &\cellcolor{lightgray!20}\textcolor{black}{\textbf{42.5}}   &
   \cellcolor{lightgray!20}\textcolor{black}{\textbf{13.3}} &
   \cellcolor{lightgray!20}\textcolor{black}{13.8}    &
   \cellcolor{lightgray!20}\textcolor{black}{19.6}     &
   \cellcolor{lightgray!20}\textcolor{black}{\textbf{186.6}} &
   \cellcolor{lightgray!20}\textcolor{black}{\textbf{488.1}}  &
   \cellcolor{lightgray!20}\textcolor{black}{50.5}    \\

   \cmidrule{2-9}
   & \cellcolor{lightgray!60}\textcolor{black}{LAUD$^\mathrm{l}$-ResNet-50 ($t$=0.5)} 
   &\cellcolor{lightgray!60}\textcolor{black}{62.2}
   &\cellcolor{lightgray!60}\textcolor{black}{15.9}
   &\cellcolor{lightgray!60}\textcolor{black}{14.5}
   &\cellcolor{lightgray!60}\textcolor{black}{21.5}
   &\cellcolor{lightgray!60}\textcolor{black}{265.8}      
   &\cellcolor{lightgray!60}\textcolor{black}{720.8}
   &\cellcolor{lightgray!60}\textcolor{black}{51.1}     \\

   & \cellcolor{lightgray!60}\textcolor{black}{LAUD$^\mathrm{l}$-ResNet-50 ($t$=0.3)}
   &\cellcolor{lightgray!60}\textcolor{black}{54.4}
   &\cellcolor{lightgray!60}\textcolor{black}{13.5}
   &\cellcolor{lightgray!60}\textcolor{black}{\textbf{12.6}}
   &\cellcolor{lightgray!60}\textcolor{black}{\textbf{18.0}}
   &\cellcolor{lightgray!60}\textcolor{black}{226.1}      
   &\cellcolor{lightgray!60}\textcolor{black}{614.7}
   &\cellcolor{lightgray!60}\textcolor{black}{50.9}     \\

   \cmidrule{2-9}

    & \textcolor{black}{ResNet-101 (Baseline)}     & \textcolor{black}{141.2} &  \textcolor{black}{33.9} & \textcolor{black}{29.8} & \textcolor{black}{44.8} & \textcolor{black}{586.4} & \textcolor{black}{1600.4}
   
    & \textcolor{black}{51.8} \\ 
   
   \cmidrule{2-9}

   & \cellcolor{lightgray!40}\textcolor{black}{LAUD$^\mathrm{s}$-ResNet-101 ($S^{\mathrm{net}}$=4-4-2-1, $t$=0.5)} & \cellcolor{lightgray!40}\textcolor{black}{93.3}  &
   \cellcolor{lightgray!40}\textcolor{black}{33.7} &
   \cellcolor{lightgray!40}\textcolor{black}{38.3}  &
   \cellcolor{lightgray!40}\textcolor{black}{41.8}  &
   \cellcolor{lightgray!40}\textcolor{black}{417.9} & 
   \cellcolor{lightgray!40}\textcolor{black}{1125.7} & 
   \cellcolor{lightgray!40}\textcolor{black}{\textbf{52.4}} \\

   & \cellcolor{lightgray!40}\textcolor{black}{LAUD$^\mathrm{s}$-ResNet-101 ($S^{\mathrm{net}}$=4-4-2-1, $t$=0.4)} & \cellcolor{lightgray!40}\textcolor{black}{85.7} &
   \cellcolor{lightgray!40}\textcolor{black}{31.5} &
   \cellcolor{lightgray!40}\textcolor{black}{35.7}  &
   \cellcolor{lightgray!40}\textcolor{black}{39.2} &
   \cellcolor{lightgray!40}\textcolor{black}{385.5}       & 
   \cellcolor{lightgray!40}\textcolor{black}{1036.7}         & 
   \cellcolor{lightgray!40}\textcolor{black}{51.9} \\

   
   \cmidrule{2-9}
   & \cellcolor{lightgray!20}\textcolor{black}{LAUD$^\mathrm{c}$-ResNet-101 ($C^{\mathrm{net}}$=2-2-2-2, $t$=0.8)} 
   &\cellcolor{lightgray!20}\textcolor{black}{111.8}   &
   \cellcolor{lightgray!20}\textcolor{black}{30.8} &
   \cellcolor{lightgray!20}\textcolor{black}{30.1}    &
   \cellcolor{lightgray!20}\textcolor{black}{42.1}     &
   \cellcolor{lightgray!20}\textcolor{black}{474.1} &
   \cellcolor{lightgray!20}\textcolor{black}{1271.8}  &
   \cellcolor{lightgray!20}\textcolor{black}{52.3}    \\


   & \cellcolor{lightgray!20}\textcolor{black}{LAUD$^\mathrm{c}$-ResNet-101 ($C^{\mathrm{net}}$=2-2-2-2, $t$=0.6)} 
   &\cellcolor{lightgray!20}\textcolor{black}{80.8}   &
   \cellcolor{lightgray!20}\textcolor{black}{24.1} &
   \cellcolor{lightgray!20}\textcolor{black}{24.7}     &
   \cellcolor{lightgray!20}\textcolor{black}{34.1}    &
   \cellcolor{lightgray!20}\textcolor{black}{338.5}  &
   \cellcolor{lightgray!20}\textcolor{black}{891.9}  &
   \cellcolor{lightgray!20}\textcolor{black}{52.0}    \\



   & \cellcolor{lightgray!20}\textcolor{black}{LAUD$^\mathrm{c}$-ResNet-101 ($C^{\mathrm{net}}$=1-1-1-1, $t$=0.5)} 
   &\cellcolor{lightgray!20}\textcolor{black}{\textbf{62.4}}   &
   \cellcolor{lightgray!20}\textcolor{black}{\textbf{20.5}} &
   \cellcolor{lightgray!20}\textcolor{black}{\textbf{21.7}}     &
   \cellcolor{lightgray!20}\textcolor{black}{\textbf{29.9}}     &
   \cellcolor{lightgray!20}\textcolor{black}{\textbf{270.2}}  &
   \cellcolor{lightgray!20}\textcolor{black}{\textbf{701.0}} &
   \cellcolor{lightgray!20}\textcolor{black}{51.8}    \\

   \cmidrule{2-9}
   & \cellcolor{lightgray!60}\textcolor{black}{LAUD$^\mathrm{l}$-ResNet-101 ($t$=0.4)} 
   &\cellcolor{lightgray!60}\textcolor{black}{104.5}
   &\cellcolor{lightgray!60}\textcolor{black}{26.0}
   &\cellcolor{lightgray!60}\textcolor{black}{23.6}
   &\cellcolor{lightgray!60}\textcolor{black}{34.4}
   &\cellcolor{lightgray!60}\textcolor{black}{439.8}      
   &\cellcolor{lightgray!60}\textcolor{black}{1197.8}
   &\cellcolor{lightgray!60}\textcolor{black}{52.2}     \\

   & \cellcolor{lightgray!60}\textcolor{black}{LAUD$^\mathrm{l}$-ResNet-101 ($t$=0.3)}
   &\cellcolor{lightgray!60}\textcolor{black}{101.9}
   &\cellcolor{lightgray!60}\textcolor{black}{25.4}
   &\cellcolor{lightgray!60}\textcolor{black}{23.1}
   &\cellcolor{lightgray!60}\textcolor{black}{33.6}
   &\cellcolor{lightgray!60}\textcolor{black}{429.0}      
   &\cellcolor{lightgray!60}\textcolor{black}{1168.3}
   &\cellcolor{lightgray!60}\textcolor{black}{51.9}     \\

  \bottomrule
  \end{tabular}
  \end{center}
  \vskip -0.15in
\end{table*}

The results are exhibited in \figurename~\ref{fig:main_results} (a). On the left we plot the relationship between accuracy and FLOPs. It becomes obvious that our LAUD-ResNets, with various granularity settings, considerably outperform competing dynamic networks. Moreover, on ResNet-101, the three paradigms seem fairly comparable, whereas, on ResNet-50, layer skipping falls behind, especially when the training target is small. This is understandable because layer skipping might be overly aggressive for more shallow models.

Interestingly, the scenario alters as we explore real latency (middle on TX2 and right on V100). On the less potent TX2, latency generally exhibits a stronger correlation with theoretical FLOPs, given that it is \emph{computation-bounded} (that means, the latency is primarily focused around computation) on such IoT devices.
However, different dynamic paradigms yield varying acceleration impacts on server-end GPU, V100, as latency could be impacted by the memory access cost. For instance, layer skipping takes precedence over the other two paradigms on the deeper ResNet-101. With the target activation rate $t\!=\!0.4$, our LAUD$^\mathrm{l}$-ResNet-101 reduces the inference latency of its static counterpart by $\sim$53\%. On the shallower ResNet-50, channel skipping keeps pace with layer skipping on some low-FLOPs models. Although our proposed course-grained spatially adaptive inference trails behind the other two schemes, it significantly outclasses the previous work using pixel-level dynamic computation \cite{verelst_dynamic_2020}. The additional results in Appendix~\ref{supp_results_IN_cls} also demonstrate the preferable efficiency of layer skipping on RTX3060 and RTX3090. Channel skipping outperforms the other two paradigms only on the edge device, Nvidia Nano.

\noindent\textbf{Lightweight baseline comparison: RegNets.}
We further evaluate our \nameShort~in lightweight CNN architectures, \emph{i.e.} RegNets-Y \cite{radosavovic2020designing}. Two different sized models are tested: RegNetY-400MF and RegNetY-800MF. Compared baselines include other types of efficient models, \emph{e.g.}, MobileNets-v2 \cite{sandler2018mobilenetv2}, ShuffletNets-v2 \cite{ma2018shufflenet} and CondenseNets \cite{huang2018condensenet}.

The results are presented in \figurename~\ref{fig:main_results} (b). We observe that while channel skipping surpasses the other two paradigms substantially in the accuracy-FLOPs trade-off, it is less efficient than layer skipping on most models except RegNet-Y-800M. Remarkably, layer skipping emerges as the most dominant paradigm. We theorize that this is due to the model width (number of channels) of RegNet-Y being limited, and the inference latency still being bounded by memory access. Moreover, layer skipping enables skipping the memory-bounded SE operation \cite{hu2018squeeze}. The results on desktop-level and server-end GPUs (Appendix~\ref{supp_results_IN_cls}) further showcase the superiority of layer skipping.

\noindent{\textcolor{black}{\textbf{Experiments on vision Transformers}. Building on the foundation laid out in Sec.~\ref{sec:arch}, our LAUDNet seamlessly integrates with vision Transformers using the AdaViT~\cite{meng2022adavit} framework. Despite the absence of direct comparisons among the three dynamic paradigms in existing studies, with \cite{meng2022adavit} employing all three simultaneously, it leaves open the question of which paradigm offers the best balance between accuracy and efficiency. We address this by showcasing the accuracy-latency trade-off curves for LAUD-T2T-ViT across various platforms—TX2, RTX3060, and V100 (the performance on RTX3090 is similar to that on V100)—in \figurename~\ref{fig:main_results} (c). The findings highlight several key insights:
\begin{itemize}
    \item Layer skipping and head (channel) skipping are more advantageous for maintaining high accuracy at high activation rates, though both experience a significant accuracy decline at reduced activation rates.
    \item When evaluating the balance between practical latency and accuracy, layer skipping \emph{consistently} outperforms head (channel) skipping on all platforms.
    \item Despite its lower theoretical upper-bound of accuracy, spatial-wise adaptive computation (token skipping) might excel over the other paradigms at lower activation rates, attributing its practical latency benefits to the straightforward implementation of indexing and selection operations on GPUs, without necessitating specialized operators as in CNNs.
    \item A synergistic application of all three paradigms further enhances the accuracy-efficiency trade-off, showing the complementary strengths of each approach.
\end{itemize}
}

\subsection{Visualization and interpretability}\label{sec_vis}

We present visualization results of \nameShort~to delve into its interpretability from the perspectives of networks’ structural redundancy and images’ spatial redundancy.

\noindent\textbf{Activation rate.} \figurename~\ref{fig_vis} (a) illustrates the average activation rates $r^\mathrm{s/c/l}$ of each block in LAUD$^\mathrm{s/c/l}$-ResNet-101 ($t$=0.5) on the ImageNet validation set. The results uncover that
\begin{itemize}
    \item The activation rate patterns for spatially dynamic convolution and layer skipping are similar. The activation rates $r^\mathrm{s}$ and $r^\mathrm{l}$ seem more binarized (close to 0 or 1) in stages 1, 2, and 4. The dynamic region/layer selection predominantly occurs in stage 3;
    \item These two paradigms tend to maintain the entire feature map ($r^\mathrm{s/l}$=1.0) at the first block of stages 2, 3, and 4, where the convolutional stride is 1. This aligns with the settings in \cite{veit2018convolutional,herrmann2018end}, where the training targets for these blocks are manually set to 1. Notably, we train our LAUDNet to meet an overall computational target, rather than confining the targets for different blocks as done in \cite{veit2018convolutional,herrmann2018end}.
    \item Channel skipping results in activation rates that are more centered around 0.5 throughout the network.
\end{itemize}

\begin{table*}
  \caption{Instance Segmentation results on the COCO dataset.}
  \vskip -0.2in
  \label{tab:coco_seg}
  \begin{center}
  \begin{tabular}{c c c c c c c c c c}
  \toprule
  Segmentation & \multirow{2}{*}{Backbone} & Backbone & \multicolumn{5}{c}{Backbone Latency (ms)} & $\mathrm{AP}^\mathrm{mask}$ & $\mathrm{AP}^\mathrm{box}$ \\
  \cmidrule{4-8}
  Framework & & FLOPs (G) & V100 & 3090 & 3060 & TX2 & Nano & (\%) & (\%) \\
  \midrule
  \multirow{9}{*}{Mask R-CNN} & ResNet-101 (Baseline)  & 141.2 &  33.9 & 29.8 & 44.8 & 586.4 & 1600.4 & 36.1 & 40.0\\ 
  \cmidrule{2-10}
  & \cellcolor{lightgray!40}LAUD$^\mathrm{s}$-ResNet-101 ($S^{\mathrm{net}}$=4-4-2-1, $t$=0.5) & \cellcolor{lightgray!40}80.5  
  & \cellcolor{lightgray!40}29.7	
  & \cellcolor{lightgray!40}33.5	
  &\cellcolor{lightgray!40}37.5	
  &\cellcolor{lightgray!40}361.9
  &\cellcolor{lightgray!40}969.9
  & \cellcolor{lightgray!40}\cellcolor{lightgray!40}\textbf{37.0} & \cellcolor{lightgray!40}\cellcolor{lightgray!40}\textbf{41.0} \\
  & \cellcolor{lightgray!40}LAUD$^\mathrm{s}$-ResNet-101 ($S^{\mathrm{net}}$=4-4-2-1, $t$=0.4) & \cellcolor{lightgray!40}\textbf{69.2}  
  & \cellcolor{lightgray!40}26.4	
  & \cellcolor{lightgray!40}29.6	
  &\cellcolor{lightgray!40}33.7	
  &\cellcolor{lightgray!40}\textbf{314.3}
  &\cellcolor{lightgray!40}\textbf{838.8}
  & \cellcolor{lightgray!40}\cellcolor{lightgray!40}36.1 & \cellcolor{lightgray!40}\cellcolor{lightgray!40}40.0 \\

  \cmidrule{2-10}
  &\cellcolor{lightgray!20}LAUD$^\mathrm{c}$-ResNet-101 ($C^{\mathrm{net}}$=2-2-2-2, $t$=0.8) 
  &\cellcolor{lightgray!20}112.7
  &\cellcolor{lightgray!20}30.7	
  &\cellcolor{lightgray!20}30.0	
  &\cellcolor{lightgray!20}42.1	
  &\cellcolor{lightgray!20}473.1
  &\cellcolor{lightgray!20}1269.3
  &\cellcolor{lightgray!20}\cellcolor{lightgray!20}36.9
  &\cellcolor{lightgray!20}\cellcolor{lightgray!20}40.9
  \\
  &\cellcolor{lightgray!20}LAUD$^\mathrm{c}$-ResNet-101 ($C^{\mathrm{net}}$=2-2-2-2, $t$=0.7) 
  &\cellcolor{lightgray!20}95.9
  &\cellcolor{lightgray!20}27.1	
  &\cellcolor{lightgray!20}27.2	
  &\cellcolor{lightgray!20}37.7	
  &\cellcolor{lightgray!20}397.8
  &\cellcolor{lightgray!20}1057.5
  &\cellcolor{lightgray!20}\cellcolor{lightgray!20}36.6
  &\cellcolor{lightgray!20}\cellcolor{lightgray!20}40.6
  \\
  &\cellcolor{lightgray!20}LAUD$^\mathrm{c}$-ResNet-101 ($C^{\mathrm{net}}$=2-2-2-2, $t$=0.6) 
  &\cellcolor{lightgray!20}80.8
  &\cellcolor{lightgray!20}23.9
  &\cellcolor{lightgray!20}24.6	
  &\cellcolor{lightgray!20}33.9	
  &\cellcolor{lightgray!20}335.7
  &\cellcolor{lightgray!20}883.9
  &\cellcolor{lightgray!20}\cellcolor{lightgray!20}36.4
  &\cellcolor{lightgray!20}\cellcolor{lightgray!20}40.3
  \\
  \cmidrule{2-10}
& \cellcolor{lightgray!60}LAUD$^\mathrm{l}$-ResNet-101 ($t$=0.5) 
   &\cellcolor{lightgray!60}101.7	
   &\cellcolor{lightgray!60}25.1			
   &\cellcolor{lightgray!60}22.9	
   &\cellcolor{lightgray!60}33.3	
  &\cellcolor{lightgray!60}424.4	
  &\cellcolor{lightgray!60}1155.7
   &\cellcolor{lightgray!60}36.7    &\cellcolor{lightgray!60}40.7 \\
& \cellcolor{lightgray!60}LAUD$^\mathrm{l}$-ResNet-101 ($t$=0.4) 
   &\cellcolor{lightgray!60}91.8	
   &\cellcolor{lightgray!60}22.8
   &\cellcolor{lightgray!60}20.9	
   &\cellcolor{lightgray!60}30.4	
   &\cellcolor{lightgray!60}384.3
   &\cellcolor{lightgray!60}1045.4
   &\cellcolor{lightgray!60}36.4    &\cellcolor{lightgray!60}40.4 \\
& \cellcolor{lightgray!60}LAUD$^\mathrm{l}$-ResNet-101 ($t$=0.3) 
   &\cellcolor{lightgray!60}82.2	
   &\cellcolor{lightgray!60}\textbf{20.6}	
   &\cellcolor{lightgray!60}\textbf{18.9}
   &\cellcolor{lightgray!60}\textbf{27.5}	
   &\cellcolor{lightgray!60}345.3
   &\cellcolor{lightgray!60}938.5
   &\cellcolor{lightgray!60}36.2    &\cellcolor{lightgray!60}40.0 \\
\midrule
\textcolor{black}{\multirow{9}{*}{Mask2Former}} & \textcolor{black}{ResNet-101 (Baseline)}  & \textcolor{black}{141.2} & \textcolor{black}{33.9} & \textcolor{black}{29.8} & \textcolor{black}{44.8} & \textcolor{black}{586.4} & \textcolor{black}{1600.4} & \textcolor{black}{44.0} & \textcolor{black}{46.7} 

\\

  \cmidrule{2-10}
  & \cellcolor{lightgray!40}\textcolor{black}{LAUD$^\mathrm{s}$-ResNet-101 ($S^{\mathrm{net}}$=4-4-2-1, $t$=0.5)} & \cellcolor{lightgray!40}\textcolor{black}{109.7}  
  & \cellcolor{lightgray!40}\textcolor{black}{37.2}	
  & \cellcolor{lightgray!40}\textcolor{black}{43.8}	
  &\cellcolor{lightgray!40}\textcolor{black}{47.1}	
  &\cellcolor{lightgray!40}\textcolor{black}{481.8}
  &\cellcolor{lightgray!40}\textcolor{black}{1301.9}
  & \cellcolor{lightgray!40}\textcolor{black}{\textbf{44.0}}
  & \cellcolor{lightgray!40}\textcolor{black}{\textbf{47.1}}
    \\
  
  & \cellcolor{lightgray!40}\textcolor{black}{LAUD$^\mathrm{s}$-ResNet-101 ($S^{\mathrm{net}}$=4-4-2-1, $t$=0.4)} & \cellcolor{lightgray!40}\textcolor{black}{98.8}  
  & \cellcolor{lightgray!40}\textcolor{black}{34.0}	
  & \cellcolor{lightgray!40}\textcolor{black}{39.9}	
  &\cellcolor{lightgray!40}\textcolor{black}{43.2}	
  &\cellcolor{lightgray!40}\textcolor{black}{436.0}
  &\cellcolor{lightgray!40}\textcolor{black}{1176.1}
  & \cellcolor{lightgray!40}\textcolor{black}{43.8}
  & \cellcolor{lightgray!40}\textcolor{black}{46.7}
  \\

  \cmidrule{2-10}
  &\cellcolor{lightgray!20}\textcolor{black}{LAUD$^\mathrm{c}$-ResNet-101 ($C^{\mathrm{net}}$=2-2-2-2, $t$=0.8)}
  &\cellcolor{lightgray!20}\textcolor{black}{111.9}
  &\cellcolor{lightgray!20}\textcolor{black}{30.5}
  &\cellcolor{lightgray!20}\textcolor{black}{29.9}
  &\cellcolor{lightgray!20}\textcolor{black}{41.8}
  &\cellcolor{lightgray!20}\textcolor{black}{469.0}
  &\cellcolor{lightgray!20}\textcolor{black}{1257.6}
  &\cellcolor{lightgray!20}\textcolor{black}{44.0}
  &\cellcolor{lightgray!20}\textcolor{black}{46.9}
  \\
  &\cellcolor{lightgray!20}\textcolor{black}{LAUD$^\mathrm{c}$-ResNet-101 ($C^{\mathrm{net}}$=2-2-2-2, $t$=0.7)}
  &\cellcolor{lightgray!20}\textcolor{black}{\textbf{94.9}}
  &\cellcolor{lightgray!20}\textcolor{black}{\textbf{26.8}}
  &\cellcolor{lightgray!20}\textcolor{black}{27.0}
  &\cellcolor{lightgray!20}\textcolor{black}{37.4}
  &\cellcolor{lightgray!20}\textcolor{black}{\textbf{393.1}}
  &\cellcolor{lightgray!20}\textcolor{black}{\textbf{1044.5}}
  &\cellcolor{lightgray!20}\textcolor{black}{43.9}
  &\cellcolor{lightgray!20}\textcolor{black}{46.8}
  \\
  \cmidrule{2-10}
& \cellcolor{lightgray!60}\textcolor{black}{LAUD$^\mathrm{l}$-ResNet-101 ($t$=0.5)} 
   &\cellcolor{lightgray!60}\textcolor{black}{112.8}	
   &\cellcolor{lightgray!60}\textcolor{black}{27.7}
   &\cellcolor{lightgray!60}\textcolor{black}{25.1}	
   &\cellcolor{lightgray!60}\textcolor{black}{36.6}	
   &\cellcolor{lightgray!60}\textcolor{black}{469.4}
   &\cellcolor{lightgray!60}\textcolor{black}{1279.2}
   &\cellcolor{lightgray!60}\textcolor{black}{44.0}    &\cellcolor{lightgray!60}\textcolor{black}{47.0} \\
& \cellcolor{lightgray!60}\textcolor{black}{LAUD$^\mathrm{l}$-ResNet-101 ($t$=0.4)} 
   &\cellcolor{lightgray!60}\textcolor{black}{109.9}	
   &\cellcolor{lightgray!60}\textcolor{black}{27.0}
   &\cellcolor{lightgray!60}\textcolor{black}{\textbf{24.5}}	
   &\cellcolor{lightgray!60}\textcolor{black}{\textbf{35.7}}	
   &\cellcolor{lightgray!60}\textcolor{black}{457.5}
   &\cellcolor{lightgray!60}\textcolor{black}{1246.6}
   &\cellcolor{lightgray!60}\textcolor{black}{43.9}    &\cellcolor{lightgray!60}\textcolor{black}{46.8} \\
  
  \bottomrule
  \end{tabular}
  \end{center}
  \vskip -0.2in
\end{table*}

\noindent\textbf{Dynamic patch selection.} We visualize the spatial masks generated by our third block of a LAUD$^\mathrm{s}$-ResNet-101 ($S^{\mathrm{net}}$=4-4-2-1) in \figurename~\ref{fig_vis} (b). The highlighted areas denote the locations of 1 elements in a mask, while computations in the dimmed regions are skipped by our dynamic model. It becomes evident that the masker is adept at pinpointing the most task-related areas, even minutiae such as the tiny aircraft at the corner, thereby trimming unnecessary computations in background zones.  Such findings imply that, a granularity of $S$=4 is aptly flexible for identifying crucial regions, paving the way for a harmonious balance between accuracy and efficiency. Intriguingly, the masker is able to pick out objects which are \emph{not labeled} for that particular sample - for instance, the flower next to the hummingbird or the person clutching the camera. This signals that our spatially dynamic networks inherently discern regions imbued with semantic significance, and their prowess isn’t shackled by mere classification labels. Such a trait is invaluable for a slew of downstream tasks, like object detection and instance segmentation (Sec.~\ref{sec_det}), tasks which necessitate the identification of various classes and objects within an image. For a broader range of visualization results, readers can refer to Appendix~\ref{sec_more_vis}.

\subsection{Dense prediction tasks}\label{sec_det}

Our \nameShort~is further put to test on downstream tasks, \emph{i.e.} COCO \cite{lin2014microsoft} object detection (as seen in Table \ref{tab:coco}) and instance segmentation (presented in Table~\ref{tab:coco_seg}). For object detection, the mean average precision (mAP) stands as the barometer for network efficacy. For instance segmentation,  the AP$^\mathrm{mask}$ dives deeper to gauge the nuance of dense prediction. The average backbone FLOPs, and the average backbone latency on the validation set are used to measure the network efficiency. \textcolor{black}{Due to LAUDNet's versatile nature, we can seamlessly replace the backbones in various detection and segmentation frameworks with our pre-trained models on ImageNet, then fine-tune them on the COCO dataset under the standard protocol for 12 epochs—except for models based on Mask2Former~\cite{cheng2022masked}, which are trained for 50 epochs in line with the baseline configurations (detailed settings are elaborated in Appendix~\ref{settings_for_det_seg}). In the domain of object detection, our experimentation covers three frameworks: the two-stage Faster R-CNN \cite{ren2015faster} with Feature Pyramid Network \cite{lin2017feature}, the one-stage RetinaNet \cite{lin2017focal}, and a DETR~\cite{carion2020end}-based model, namely Dense Distinct Query (DDQ)-DETR~\cite{zhang2023dense}. We compare our results against a range of recent advancements, such as Deformable DETR~\cite{zhu2020deformable}, DINO-DETR~\cite{zhang2022dino}, Rank-DETR~\cite{pu2023rank}, and Stable-DINO~\cite{liu2023detection}. For instance segmentation, we utilize the well-established Mask R-CNN \cite{he2017mask} and the query-based Mask2Former~\cite{cheng2022masked}. The results are presented in Tab.~\ref{tab:coco} (for object detection) and Tab.~\ref{tab:coco_seg} (for instance segmentation), unequivocally demonstrating that LAUDNet consistently boosts both mAP and efficiency across classic and state-of-the-art (SOTA) frameworks. Notably, while channel and layer skipping generally surpass spatial-wise dynamic computation in efficiency, the ideal dynamic paradigm may vary depending on the specific detection framework, backbone architecture, and hardware platforms.}

\section{Conclusion}
In this paper, we propose to build \emph{latency-aware} unified dynamic networks (\nameShort) under the guidance of a \emph{latency prediction model}. By collectively considering the algorithm, scheduling strategy, and hardware properties, we can accurately estimate the practical latency of different dynamic operators on any computing platforms. Based on an empirical analysis of the correlation between latency and the \emph{granularity} of spatial-wise and channel-wise adaptive inference, the algorithm and scheduling strategies are optimized to attain realistic speedup on a range of multi-core processors, such as Tesla V100 and Jetson TX2. Our experiments on image classification, object detection, and instance segmentation tasks affirm that the proposed method markedly boosts the practical efficiency of deep CNNs and surpasses numerous competing approaches. We believe our research brings useful insights into the design of dynamic networks. Future works include explorations on more types of model architectures (\emph{e.g.} Transformers, large language models) and tasks (\emph{e.g.} low-level vision tasks and vision-language tasks).

\ifCLASSOPTIONcompsoc
  \section*{Acknowledgments}
\else
  \section*{Acknowledgment}
\fi

This work is supported in part by the National Key R\&D Program of China under Grant 2021ZD0140407, the National Natural Science Foundation of China under Grants 42327901 and 62276150, and Guoqiang Institute of Tsinghua University. We also appreciate the generous donation of computing resources by High-Flyer AI.


\small{
    \bibliography{ref}

\begin{thebibliography}{10}

\bibitem{deng2009imagenet}
Jia Deng, Wei Dong, Richard Socher, Li-Jia Li, Kai Li, and Li~Fei-Fei.
\newblock Imagenet: A large-scale hierarchical image database.
\newblock In {\em CVPR}, 2009.

\bibitem{he2016resnet}
Kaiming He, Xiangyu Zhang, Shaoqing Ren, and Jian Sun.
\newblock Deep residual learning for image recognition.
\newblock In {\em CVPR}, 2016.

\bibitem{huang2017densely}
Gao Huang, Zhuang Liu, Geoff Pleiss, Laurens Van Der~Maaten, and Kilian
  Weinberger.
\newblock Densely connected convolutional networks.
\newblock {\em TPAMI}, 2019.

\bibitem{dosovitskiy2020image}
Alexey Dosovitskiy, Lucas Beyer, Alexander Kolesnikov, Dirk Weissenborn,
  Xiaohua Zhai, Thomas Unterthiner, Mostafa Dehghani, Matthias Minderer, Georg
  Heigold, Sylvain Gelly, Jakob Uszkoreit, and Neil Houlsby.
\newblock An image is worth 16x16 words: Transformers for image recognition at
  scale.
\newblock In {\em ICLR}, 2021.

\bibitem{kirillov2023segment}
Alexander Kirillov, Eric Mintun, Nikhila Ravi, Hanzi Mao, Chloe Rolland, Laura
  Gustafson, Tete Xiao, Spencer Whitehead, Alexander~C Berg, Wan-Yen Lo, et~al.
\newblock Segment anything.
\newblock {\em arXiv preprint arXiv:2304.02643}, 2023.

\bibitem{vaswani2017attention}
Ashish Vaswani, Noam Shazeer, Niki Parmar, Jakob Uszkoreit, Llion Jones,
  Aidan~N Gomez, {\L}ukasz Kaiser, and Illia Polosukhin.
\newblock Attention is all you need.
\newblock In {\em NeurIPS}, 2017.

\bibitem{devlin_bert_2019}
Jacob Devlin, Ming-Wei Chang, Kenton Lee, and Kristina Toutanova.
\newblock {BERT}: {Pre}-training of {Deep} {Bidirectional} {Transformers} for
  {Language} {Understanding}.
\newblock In {\em ACL}, 2019.

\bibitem{radford2018improving}
Alec Radford, Karthik Narasimhan, Tim Salimans, Ilya Sutskever, et~al.
\newblock Improving language understanding by generative pre-training.
\newblock 2018.

\bibitem{radford2019language}
Alec Radford, Jeffrey Wu, Rewon Child, David Luan, Dario Amodei, and Ilya
  Sutskever.
\newblock Language models are unsupervised multitask learners.
\newblock {\em OpenAI blog}, 2019.

\bibitem{openai2023gpt4}
OpenAI.
\newblock Gpt-4 technical report, 2023.

\bibitem{han2021dynamic}
Yizeng Han, Gao Huang, Shiji Song, Le~Yang, Honghui Wang, and Yulin Wang.
\newblock Dynamic neural networks: A survey.
\newblock {\em TPAMI}, 2021.

\bibitem{huang2017multi}
Gao Huang, Danlu Chen, Tianhong Li, Felix Wu, Laurens van~der Maaten, and
  Kilian Weinberger.
\newblock Multi-scale dense networks for resource efficient image
  classification.
\newblock In {\em ICLR}, 2018.

\bibitem{han2022learning}
Yizeng Han, Yifan Pu, Zihang Lai, Chaofei Wang, Shiji Song, Junfen Cao, Wenhui
  Huang, Chao Deng, and Gao Huang.
\newblock Learning to weight samples for dynamic early-exiting networks.
\newblock In {\em ECCV}, 2022.

\bibitem{wang2018skipnet}
Xin Wang, Fisher Yu, Zi-Yi Dou, Trevor Darrell, and Joseph~E Gonzalez.
\newblock Skipnet: Learning dynamic routing in convolutional networks.
\newblock In {\em ECCV}, 2018.

\bibitem{veit2018convolutional}
Andreas Veit and Serge Belongie.
\newblock Convolutional networks with adaptive inference graphs.
\newblock In {\em ECCV}, 2018.

\bibitem{lin2017runtime}
Ji~Lin, Yongming Rao, Jiwen Lu, and Jie Zhou.
\newblock Runtime neural pruning.
\newblock In {\em NeurIPS}, 2017.

\bibitem{bejnordi2019batch}
Babak~Ehteshami Bejnordi, Tijmen Blankevoort, and Max Welling.
\newblock Batch-shaping for learning conditional channel gated networks.
\newblock In {\em ICLR}, 2020.

\bibitem{figurnov2017spatially}
Michael Figurnov, Maxwell~D Collins, Yukun Zhu, Li~Zhang, Jonathan Huang,
  Dmitry Vetrov, and Ruslan Salakhutdinov.
\newblock Spatially adaptive computation time for residual networks.
\newblock In {\em CVPR}, 2017.

\bibitem{dong_more_2017}
Xuanyi Dong, Junshi Huang, Yi~Yang, and Shuicheng Yan.
\newblock More is less: {A} more complicated network with less inference
  complexity.
\newblock In {\em CVPR}, 2017.

\bibitem{verelst_dynamic_2020}
Thomas Verelst and Tinne Tuytelaars.
\newblock Dynamic convolutions: {Exploiting} {Spatial} {Sparsity} for {Faster}
  {Inference}.
\newblock In {\em CVPR}, 2020.

\bibitem{xie2020spatially}
Zhenda Xie, Zheng Zhang, Xizhou Zhu, Gao Huang, and Stephen Lin.
\newblock Spatially adaptive inference with stochastic feature sampling and
  interpolation.
\newblock In {\em ECCV}, 2020.

\bibitem{wang2020glance}
Yulin Wang, Kangchen Lv, Rui Huang, Shiji Song, Le~Yang, and Gao Huang.
\newblock Glance and focus: a dynamic approach to reducing spatial redundancy
  in image classification.
\newblock In {\em NeurIPS}, 2020.

\bibitem{SAR_TIP}
Yizeng Han, Gao Huang, Shiji Song, Le~Yang, Yitian Zhang, and Haojun Jiang.
\newblock Spatially adaptive feature refinement for efficient inference.
\newblock {\em TIP}, 2021.

\bibitem{radosavovic2020designing}
Ilija Radosavovic, Raj~Prateek Kosaraju, Ross Girshick, Kaiming He, and Piotr
  Doll{\'a}r.
\newblock Designing network design spaces.
\newblock In {\em CVPR}, 2020.

\bibitem{touvron2021training}
Hugo Touvron, Matthieu Cord, Matthijs Douze, Francisco Massa, Alexandre
  Sablayrolles, and Herv{\'e} J{\'e}gou.
\newblock Training data-efficient image transformers \& distillation through
  attention.
\newblock In {\em ICML}, 2021.

\bibitem{yuan2021tokens}
Li~Yuan, Yunpeng Chen, Tao Wang, Weihao Yu, Yujun Shi, Zi-Hang Jiang,
  Francis~EH Tay, Jiashi Feng, and Shuicheng Yan.
\newblock Tokens-to-token vit: Training vision transformers from scratch on
  imagenet.
\newblock In {\em ICCV}, 2021.

\bibitem{han2022latency}
Yizeng Han, Zhihang Yuan, Yifan Pu, Chenhao Xue, Shiji Song, Guangyu Sun, and
  Gao Huang.
\newblock Latency-aware spatial-wise dynamic networks.
\newblock In {\em NeurIPS}, 2022.

\bibitem{cutlass}
Nvidia.
\newblock Cutlass.
\newblock \url{https://www.github.com/NVIDIA/cutlass}.

\bibitem{howard2017mobilenets}
Andrew~G Howard, Menglong Zhu, Bo~Chen, Dmitry Kalenichenko, Weijun Wang,
  Tobias Weyand, Marco Andreetto, and Hartwig Adam.
\newblock Mobilenets: Efficient convolutional neural networks for mobile vision
  applications.
\newblock {\em arXiv preprint arXiv:1704.04861}, 2017.

\bibitem{sandler2018mobilenetv2}
Mark Sandler, Andrew Howard, Menglong Zhu, Andrey Zhmoginov, and Liang-Chieh
  Chen.
\newblock Mobilenetv2: Inverted residuals and linear bottlenecks.
\newblock In {\em CVPR}, 2018.

\bibitem{zhang2018shufflenet}
Xiangyu Zhang, Xinyu Zhou, Mengxiao Lin, and Jian Sun.
\newblock Shufflenet: An extremely efficient convolutional neural network for
  mobile devices.
\newblock In {\em CVPR}, 2018.

\bibitem{ma2018shufflenet}
Ningning Ma, Xiangyu Zhang, Hai-Tao Zheng, and Jian Sun.
\newblock Shufflenet v2: Practical guidelines for efficient cnn architecture
  design.
\newblock In {\em ECCV}, 2018.

\bibitem{han2015deepcompression}
Song Han, Huizi Mao, and William~J Dally.
\newblock Deep compression: Compressing deep neural networks with pruning,
  trained quantization and huffman coding.
\newblock 2016.

\bibitem{he2018pruning}
Yang He, Ping Liu, Ziwei Wang, Zhilan Hu, and Yi~Yang.
\newblock Filter pruning via geometric median for deep convolutional neural
  networks acceleration.
\newblock In {\em CVPR}, 2019.

\bibitem{huang2018condensenet}
Gao Huang, Shichen Liu, Laurens Van~der Maaten, and Kilian~Q Weinberger.
\newblock Condensenet: An efficient densenet using learned group convolutions.
\newblock In {\em CVPR}, 2018.

\bibitem{yang2021condensenet}
Le~Yang, Haojun Jiang, Ruojin Cai, Yulin Wang, Shiji Song, Gao Huang, and
  Qi~Tian.
\newblock Condensenet v2: Sparse feature reactivation for deep networks.
\newblock In {\em CVPR}, 2021.

\bibitem{hubara2016binarized}
Itay Hubara, Matthieu Courbariaux, Daniel Soudry, Ran El-Yaniv, and Yoshua
  Bengio.
\newblock Binarized neural networks.
\newblock In {\em NeurIPS}, 2016.

\bibitem{choi2018pact}
Jungwook Choi, Zhuo Wang, Swagath Venkataramani, Pierce I-Jen Chuang,
  Vijayalakshmi Srinivasan, and Kailash Gopalakrishnan.
\newblock Pact: Parameterized clipping activation for quantized neural
  networks.
\newblock {\em arXiv preprint arXiv:1805.06085}, 2018.

\bibitem{jung2019learning}
Sangil Jung, Changyong Son, Seohyung Lee, Jinwoo Son, Jae-Joon Han, Youngjun
  Kwak, Sung~Ju Hwang, and Changkyu Choi.
\newblock Learning to quantize deep networks by optimizing quantization
  intervals with task loss.
\newblock In {\em CVPR}, 2019.

\bibitem{yuan2022ptq4vit}
Zhihang Yuan, Chenhao Xue, Yiqi Chen, Qiang Wu, and Guangyu Sun.
\newblock Ptq4vit: Post-training quantization for vision transformers with twin
  uniform quantization.
\newblock In {\em ECCV}, 2022.

\bibitem{hinton2014distilling}
Geoffrey Hinton, Oriol Vinyals, and Jeff Dean.
\newblock Distilling the knowledge in a neural network.
\newblock In {\em NeurIPS Workshop}, 2014.

\bibitem{wang2022efficient}
Chaofei Wang, Qisen Yang, Rui Huang, Shiji Song, and Gao Huang.
\newblock Efficient knowledge distillation from model checkpoints.
\newblock In {\em NeurIPS}, 2022.

\bibitem{graves2016adaptive}
Alex Graves.
\newblock Adaptive computation time for recurrent neural networks.
\newblock {\em arXiv preprint arXiv:1603.08983}, 2016.

\bibitem{yang2020resolution}
Le~Yang, Yizeng Han, Xi~Chen, Shiji Song, Jifeng Dai, and Gao Huang.
\newblock Resolution adaptive networks for efficient inference.
\newblock In {\em CVPR}, 2020.

\bibitem{gao2018dynamic}
Xitong Gao, Yiren Zhao, Łukasz Dudziak, Robert Mullins, and Cheng zhong Xu.
\newblock Dynamic channel pruning: Feature boosting and suppression.
\newblock In {\em ICLR}, 2019.

\bibitem{cai2019proxylessnas}
Han Cai, Ligeng Zhu, and Song Han.
\newblock Proxylessnas: Direct neural architecture search on target task and
  hardware.
\newblock In {\em ICLR}, 2019.

\bibitem{tan2019mnasnet}
Mingxing Tan, Bo~Chen, Ruoming Pang, Vijay Vasudevan, Mark Sandler, Andrew
  Howard, and Quoc~V Le.
\newblock Mnasnet: Platform-aware neural architecture search for mobile.
\newblock In {\em Proceedings of the IEEE/CVF Conference on Computer Vision and
  Pattern Recognition}, pages 2820--2828, 2019.

\bibitem{wu2019fbnet}
Bichen Wu, Xiaoliang Dai, Peizhao Zhang, Yanghan Wang, Fei Sun, Yiming Wu,
  Yuandong Tian, Peter Vajda, Yangqing Jia, and Kurt Keutzer.
\newblock Fbnet: Hardware-aware efficient convnet design via differentiable
  neural architecture search.
\newblock In {\em CVPR}, 2019.

\bibitem{cai2019once}
Han Cai, Chuang Gan, Tianzhe Wang, Zhekai Zhang, and Song Han.
\newblock Once-for-all: Train one network and specialize it for efficient
  deployment.
\newblock In {\em ICLR}, 2019.

\bibitem{zoph2016neural}
Barret Zoph and Quoc~V Le.
\newblock Neural architecture search with reinforcement learning.
\newblock In {\em ICLR}, 2017.

\bibitem{liu_darts_2018}
Hanxiao Liu, Karen Simonyan, and Yiming Yang.
\newblock {DARTS}: {Differentiable} {Architecture} {Search}.
\newblock In {\em ICLR}, 2018.

\bibitem{ren_sbnet_2018}
Mengye Ren, Andrei Pokrovsky, Bin Yang, and Raquel Urtasun.
\newblock {SBNet}: {Sparse} {Blocks} {Network} for {Fast} {Inference}.
\newblock {\em CVPR}, 2018.

\bibitem{wu2018blockdrop}
Zuxuan Wu, Tushar Nagarajan, Abhishek Kumar, Steven Rennie, Larry~S Davis,
  Kristen Grauman, and Rogerio Feris.
\newblock Blockdrop: Dynamic inference paths in residual networks.
\newblock In {\em CVPR}, 2018.

\bibitem{herrmann2018end}
Charles Herrmann, Richard~Strong Bowen, and Ramin Zabih.
\newblock Channel selection using gumbel softmax.
\newblock In {\em ECCV}, 2020.

\bibitem{jang2016categorical}
Eric Jang, Shixiang Gu, and Ben Poole.
\newblock Categorical reparameterization with gumbel-softmax.
\newblock In {\em ICLR}, 2017.

\bibitem{maddison2016concrete}
Chris~J Maddison, Andriy Mnih, and Yee~Whye Teh.
\newblock The concrete distribution: A continuous relaxation of discrete random
  variables.
\newblock In {\em ICLR}, 2017.

\bibitem{meng2022adavit}
Lingchen Meng, Hengduo Li, Bor-Chun Chen, Shiyi Lan, Zuxuan Wu, Yu-Gang Jiang,
  and Ser-Nam Lim.
\newblock Adavit: Adaptive vision transformers for efficient image recognition.
\newblock In {\em CVPR}, 2022.

\bibitem{hennessy2011computer}
John~L Hennessy and David~A Patterson.
\newblock {\em Computer architecture: a quantitative approach}.
\newblock Elsevier, 2011.

\bibitem{hu2018squeeze}
Jie Hu, Li~Shen, and Gang Sun.
\newblock Squeeze-and-excitation networks.
\newblock In {\em CVPR}, 2018.

\bibitem{colleman2021processor}
Steven Colleman, Thomas Verelst, Linyan Mei, Tinne Tuytelaars, and Marian
  Verhelst.
\newblock Processor architecture optimization for spatially dynamic neural
  networks.
\newblock In {\em VLSI-SoC}, 2021.

\bibitem{lin2014microsoft}
Tsung-Yi Lin, Michael Maire, Serge Belongie, James Hays, Pietro Perona, Deva
  Ramanan, Piotr Doll{\'a}r, and C~Lawrence Zitnick.
\newblock Microsoft coco: Common objects in context.
\newblock In {\em ECCV}, 2014.

\bibitem{cheng2022masked}
Bowen Cheng, Ishan Misra, Alexander~G Schwing, Alexander Kirillov, and Rohit
  Girdhar.
\newblock Masked-attention mask transformer for universal image segmentation.
\newblock In {\em CVPR}, 2022.

\bibitem{ren2015faster}
Shaoqing Ren, Kaiming He, Ross Girshick, and Jian Sun.
\newblock Faster r-cnn: Towards real-time object detection with region proposal
  networks.
\newblock In {\em NeurIPS}, 2015.

\bibitem{lin2017feature}
Tsung-Yi Lin, Piotr Doll{\'a}r, Ross Girshick, Kaiming He, Bharath Hariharan,
  and Serge Belongie.
\newblock Feature pyramid networks for object detection.
\newblock In {\em CVPR}, 2017.

\bibitem{lin2017focal}
Tsung-Yi Lin, Priya Goyal, Ross Girshick, Kaiming He, and Piotr Doll{\'a}r.
\newblock Focal loss for dense object detection.
\newblock In {\em ICCV}, 2017.

\bibitem{carion2020end}
Nicolas Carion, Francisco Massa, Gabriel Synnaeve, Nicolas Usunier, Alexander
  Kirillov, and Sergey Zagoruyko.
\newblock End-to-end object detection with transformers.
\newblock In {\em ECCV}, 2020.

\bibitem{zhang2023dense}
Shilong Zhang, Xinjiang Wang, Jiaqi Wang, Jiangmiao Pang, Chengqi Lyu, Wenwei
  Zhang, Ping Luo, and Kai Chen.
\newblock Dense distinct query for end-to-end object detection.
\newblock In {\em CVPR}, 2023.

\bibitem{zhu2020deformable}
Xizhou Zhu, Weijie Su, Lewei Lu, Bin Li, Xiaogang Wang, and Jifeng Dai.
\newblock Deformable detr: Deformable transformers for end-to-end object
  detection.
\newblock In {\em ICLR}, 2020.

\bibitem{zhang2022dino}
Hao Zhang, Feng Li, Shilong Liu, Lei Zhang, Hang Su, Jun Zhu, Lionel Ni, and
  Heung-Yeung Shum.
\newblock Dino: Detr with improved denoising anchor boxes for end-to-end object
  detection.
\newblock In {\em ICLR}, 2022.

\bibitem{pu2023rank}
Yifan Pu, Weicong Liang, Yiduo Hao, Yuhui Yuan, Yukang Yang, Chao Zhang, Han
  Hu, and Gao Huang.
\newblock Rank-detr for high quality object detection.
\newblock In {\em NeurIPS}, 2023.

\bibitem{liu2023detection}
Shilong Liu, Tianhe Ren, Jiayu Chen, Zhaoyang Zeng, Hao Zhang, Feng Li,
  Hongyang Li, Jun Huang, Hang Su, Jun Zhu, et~al.
\newblock Detection transformer with stable matching.
\newblock {\em arXiv preprint arXiv:2304.04742}, 2023.

\bibitem{he2017mask}
Kaiming He, Georgia Gkioxari, Piotr Doll{\'a}r, and Ross Girshick.
\newblock Mask r-cnn.
\newblock In {\em ICCV}, pages 2961--2969, 2017.

\end{thebibliography}
}


%



\ifCLASSOPTIONcaptionsoff
  \newpage
\fi

\appendices
\section{Latency prediction model.}\label{detailed_latency_predict} 
As the dynamic operators in our method have not been supported by current deep learning libraries, we propose a latency prediction model to efficiently estimate the real latency of these operators on hardware device. The inputs of the latency prediction model include: 1) the structural configuration and dynamic paradigm of a convolution block, 2) its activation rate $r$ which decides the computation amount, 3) the spatial (channel) granularity $S$ ($G$), and 4) the hardware properties mentioned in Table~\ref{tab_hardware_property}. The latency of a dynamic block is predicted as follows.

\textbf{Input/output shape definition}. The first step of predicting the latency of an operation is to calculate the shape of input and output. 
Taking the gather-conv2 operation in LAUD$^\mathrm{s}$-ResNets as an example, the input of this operation is the activation with the shape of $C_\mathrm{in} \!\times\!H\!\times\! W $, where $C_\mathrm{in}$ is the number of input channels, and $H$ and $W$ are the resolution of the feature map.
The shape of the output tensor is  $ P \times C_\mathrm{out}\!\times\! S \!\times\! S $, where $P$ is the number of output patches, $C_\mathrm{out}$ is the number of output channels and $S$ is the spatial granularity. Note that $P$ is obtained based on the output of our maskers.

\textbf{Operation-to-hardware mapping.} Next, we map the operations to hardware.
As illustrated in \figurename~\ref{hardware_model}, we model a hardware device as multiple processing engines (PEs). 
We assign the computation of each element in the output feature map to a PE. 
Specifically, we consecutively split the output feature map into multiple \emph{tiles}. 
The shape of each tile is $T_P \times T_C \times T_{S1} \times T_{S2}$.
These split tiles are assigned to multiple PEs.
The computation of the elements in each tile is executed in a PE. 
We can configure different shapes of tiles. 
In order to determine the optimal shape of the tile, we make a search space of different tile shapes.
The tile shape has 4 dimensions.
The candidates of each dimension are power-of-2 and do not exceed the corresponding dimension of the feature map.

\textbf{Latency estimation.} Then, we evaluate the latency of each tile shape in the search space and select the optimal tile shape with the lowest latency. 
The latency includes the \emph{data movement} latency and the \emph{computation} latency: 
\begin{equation}
\ell=\ell_{\mathrm{data}}+\ell_{\mathrm{computation}}.
\end{equation}

1) \emph{Data movement latency $\ell_\mathrm{data}$.} The estimation of the latency for data movement requires us to model the memory system of a hardware device. We model the memory system of hardware as a three-level architecture \cite{hennessy2011computer}: off-chip memory, on-chip global memory, and local memory in PE.
The input data and weight data are first transferred from the off-chip memory to the on-chip global memory. 
We assume the hardware can make full use of the off-chip memory bandwidth to simplify the latency prediction model.

After that, the data used to compute the output tiles is transferred from on-chip global memory to the local memory of each PE. 
The latency of data movement to local memory is estimated by its \emph{bandwidth} and \emph{efficiency}. 
We assume each PE only moves the corresponding input feature maps and weights once to compute a output tile so as to simplify the prediction model.
The input data movement latency $\ell_\mathrm{in}$ is calculated by adding the time from off-chip memory to on-chip global memory and the time from on-chip global memory to local-memory together: $\ell_\mathrm{in}=\ell_\mathrm{off2on} + \ell_\mathrm{global2local}$.
Contrary to the input data, the output data $\ell_\mathrm{out}$ are moved from local memory to on-chip global memory and then to off-chip memory: $\ell_\mathrm{out}=\ell_\mathrm{local2global} + \ell_\mathrm{on2off}$.
We calculate the total data movement latency by adding the input and output data movement latency together: 
\begin{equation}
 \ell_\mathrm{data}=\ell_\mathrm{in} + \ell_\mathrm{out}.   
\end{equation}
\begin{table}
  \caption{Hardware properties.}
  \vskip -0.1in
  \label{tab_hardware_property}
  \centering
  \resizebox{\linewidth}{!}{
  \begin{tabular}{ccccc}
    \toprule
    Name & \#PE     &    \#FP32  &   Frequency   & Bandwidth \\
    \midrule
    Nvidia Tesla V100    & 80 & 64  & 1500MHz  & 700G  \\
    Nvidia RTX3090       & 82 & 128 & 1695MHz  & 936G  \\
    Nvidia RTX3060       & 28 & 128  & 1777MHz  & 360G  \\
    Nvidia Jetson TX2    & 2  & 128 & 1300MHz  & 59.7G \\
    Nvidia Nano          & 1  & 128 & 921MHz   & 25.6G \\
    \bottomrule
  \end{tabular}
  }
\end{table}
\begin{figure} 
    \centering
    \includegraphics[width=0.85\linewidth]{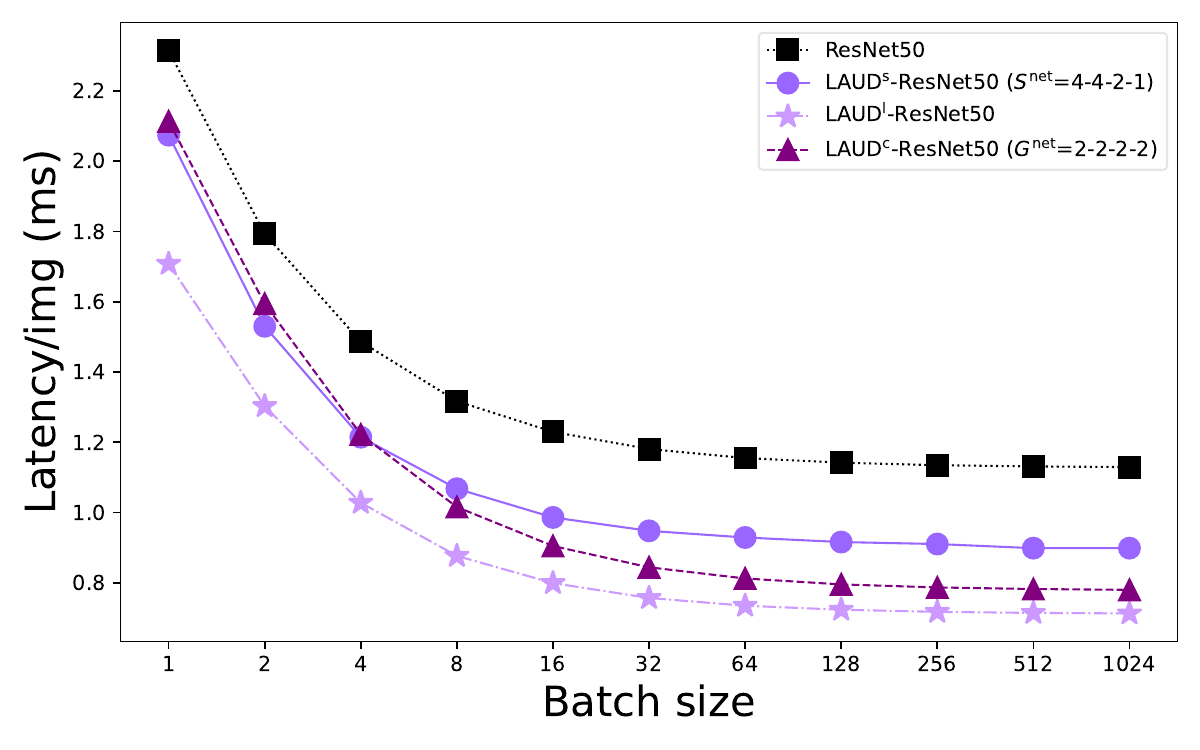} 
    \vskip -0.05in
    \caption{Relationship between the inference latency per image and batch size of LAUD-ResNet-50 on RTX3060.} 
    \label{fig:latency_bs_3060}
    \vskip -0.1in
\end{figure}

\begin{table*}
  \caption{Ablation study of channel masker.}
  \vskip -0.1in
  \label{tab_channel_masker}
  \centering
  \begin{tabular}{c|cc|cc|cc|cc}
    \toprule
    
    \multirow{2}{*}{Stage} & \multicolumn{2}{c}{1} & \multicolumn{2}{c}{2} & \multicolumn{2}{c}{3} & \multicolumn{2}{c}{4} \\
     & Latency (ms) & Ratio & Latency (ms) & Ratio & Latency (ms) & Ratio & Latency (ms) & Ratio \\
    \midrule
    1-layer & 0.21 & 14.9\% & 0.29 & 16.5\% & 0.58 & 21.7\% & 0.72 & 33.6\% \\
    \cellcolor{gray!30}2-layer (ours) & \cellcolor{gray!30}0.22 & \cellcolor{gray!30}15.8\% & \cellcolor{gray!30}0.27 & \cellcolor{gray!30}15.6\% & \cellcolor{gray!30}0.31 & \cellcolor{gray!30}12.8\% & \cellcolor{gray!30}0.19 & \cellcolor{gray!30}11.6\% \\
    \bottomrule
  \end{tabular}
\end{table*}

\begin{figure*}
     \centering
     \includegraphics[width=\linewidth]{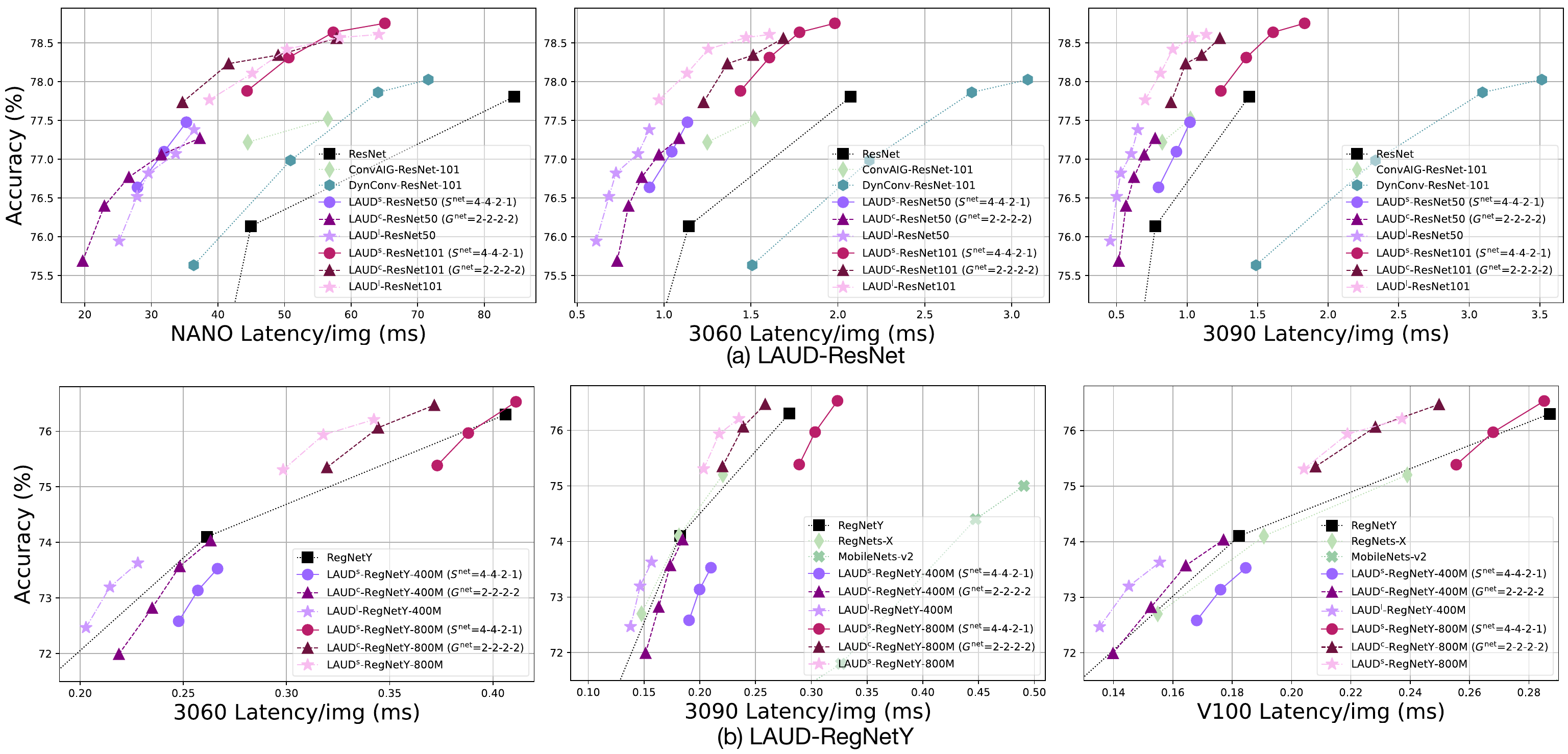}
     \vskip -0.1in
    \caption{Additional results on more hardware devices. (a) for ResNets and (b) for RegNets.}
     \label{fig_3060_nano}
     \vskip -0.2in
\end{figure*}

\begin{figure*}
     \centering
     \centering
     \includegraphics[width=\linewidth]{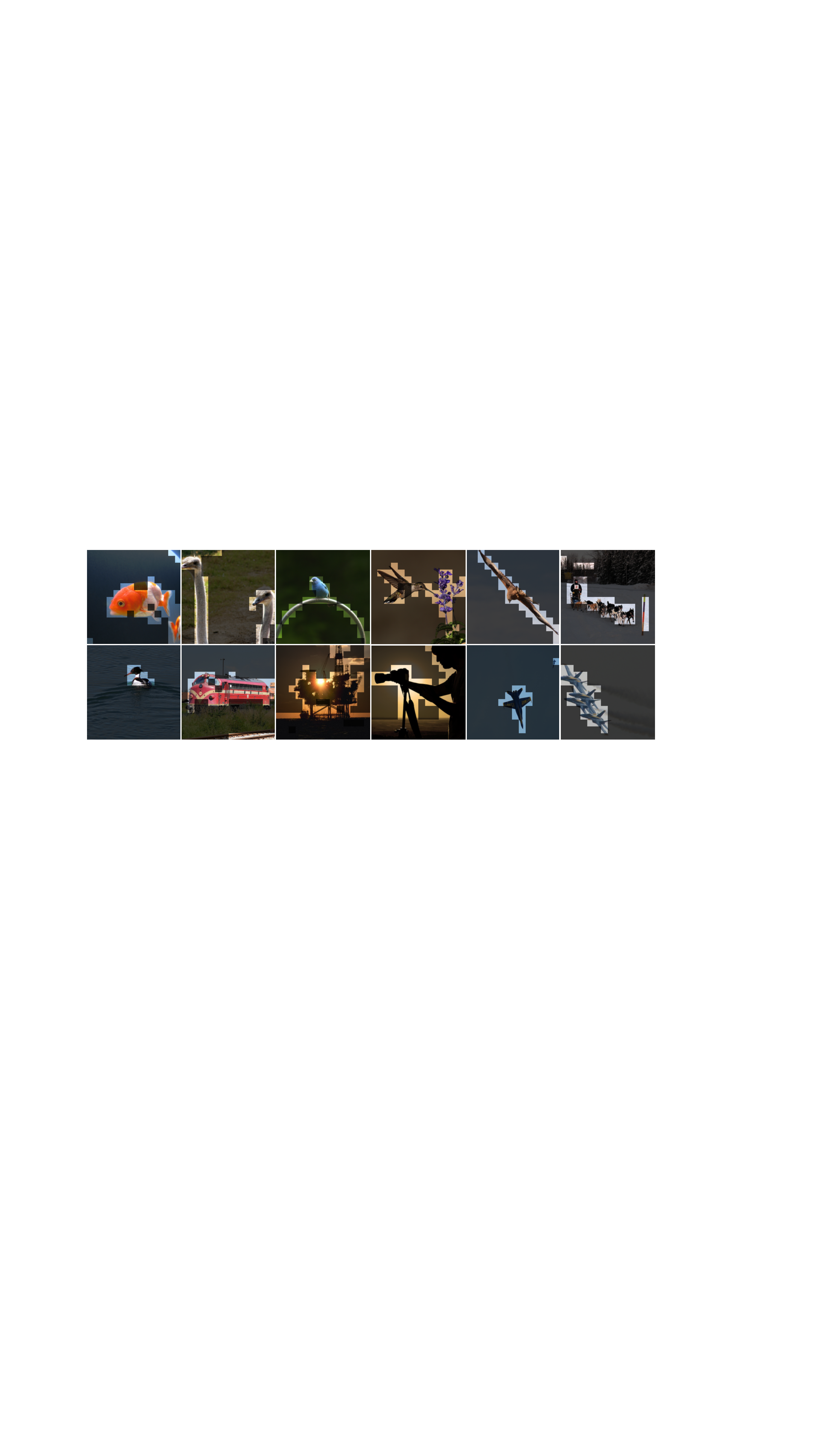}
     \vskip -0.1in
     \caption{Additional visualization results of LAUD$^\mathrm{s}$-ResNet-101.}
     \vskip -0.2in
     \label{fig_vis_supp}
\end{figure*}

The latency of data movement is affected by the granularity $S$ or $G$: when the granularity is small, the same input data has a higher probability of being sent to multiple PEs to compute different output patches, which significantly increases the number of on-chip memory movement.
And due to the small amount of data transmitted each time and the data is randomly distributed, the efficiency of data movement will be low. 
This accounts for our experiment results in the paper that \emph{a larger $S$ will effectively improve the practical efficiency}.

2) \emph{Computation latency $\ell_\mathrm{computation}$.} The computation latency of each tile is estimated using the PE's \emph{maximum throughput of FP32 computation} and the \emph{FLOPs} of computing an output tile. The total computation latency can be obtained according to the number of tiles and the number of PEs. 

To summarize, our latency prediction model can predict the real latency of dynamic operators by considering both the \emph{data movement} cost and the \emph{computation} cost.
Guided by the latency prediction model, we propose our \nameShort~with coarse-grained spatially adaptive inference ($S\!>\!1$ and $G\!>\!1$).  It is validated in our paper that \nameShort~achieve better efficiency than previous approaches \cite{xie2020spatially, verelst_dynamic_2020} ($S\!=\!1$), as it effectively reduces the data movement latency, which is rarely considered by other researchers.

\section{Detailed experimental settings}\label{sec_detailed_settings}
In this section, we present the detailed experiment settings which are not provided in the main paper due to the page limit.
\subsection{Latency prediction}\label{sec_detailed_settings_for_latency_predict}

\noindent\textbf{Hardware properties} considered by our latency prediction model include the number of processing engines (\#PE), the floating-point computation in a processing engine (\#FP32), the frequency and the bandwidth. We test four types of hardware devices, and their properties are listed in Table~\ref{tab_hardware_property}.

It could be found that the server-end GPUs V100 and RTX3090 are more powerful hardware devices, especially with the largest number of processing engines (\#PE). Therefore, spatially adaptive inference and dynamic channel skipping could easily fall into a \emph{memory-bounded} operation on these GPUs. Our experiment results in \figurename~\ref{fig:r_l_vs_r_s_and_S} and \figurename~\ref{fig:compare_S_G} in the paper can reflect this phenomenon: the more flexibility the computation is, the harder to improve the practical efficiency. 


\noindent\textbf{Operator fusion.} 

1) \emph{Fusing the masker and the first convolution.} We mentioned in Sec.~\ref{sec_schedule_optim} of the paper that the masker operation is fused with the first 1$\times$1 convolution in a block to reduce the cost on memory access. This is feasible because the two operators share the same input feature, and their convolution kernel sizes are both 1$\times$1. 

Note that during the inference stage, we only need to perform $\arg\max$ along the channel dimension of a mask $\mathbf{M}\in\mathbb{R}^{2\times H\times W}$ to obtain the positions of the gathered pixels. Therefore, we can reduce the output channel number of our maskers from 2 to 1 since the convolution is a linear operation: 
\begin{equation}
 [\mathbf{x}*\mathbf{W}]_{:,:,0}>[\mathbf{x}*\mathbf{W}]_{:,:,1} \Longleftrightarrow \mathbf{x}*(\mathbf{W}_{:,:,0}-\mathbf{W}_{:,:,1}) > 0.
\end{equation}

Afterwards, we fuse the masker with the first convolution by performing once convolution whose output channel number is $C+1$, where $C$ is the original output width of the first convolution. The output of this step is split into a feature map (for further computation) and a mask (for obtaining the index for gathering). Such operator fusion avoids the repeated reading the input feature, and helps reduce the inference latency (Tab.~\ref{ablation_op_fusion}).

2) \emph{Fusing the gather operation and the dynamic convolution.}
To facilitate the scheduling on hardware devices with multiple PEs, the masker generates the indices of activated patches instead of sparse mask at inference time. 
In this way, it is easy to evenly distribute the computation of output patches to different PEs, thus avoiding unbalanced computation of PEs.
Each element in the indices represents the index of an activated patch. 
PE fetches the input data from the corresponding positions on the feature map according to the index.
The output patches could be densely stored in memory. Such operator fusion benefits the contiguous memory access and parallel computation on multiple PEs.

3) \emph{Fusing the scatter operation and the add operation.} Similar to the previous operation, each PE fetches a tile of data from the residual feature map according to the index, adds them with the corresponding feature map from previous dynamic convolution, and then stores the results to the corresponding position on the residual feature map according to the index.
This optimization can significantly reduce the costs on memory access.

\noindent\textbf{Speed test.} We test the latency on real hardware devices to evaluate the accuracy of our latency prediction model.
On GPUs, we use Nvidia Cutlass (\url{https://github.com/NVIDIA/cutlass}) and CUDA (version 11.6) for code generation and compilation respectively. The results in \figurename~\ref{real_predicted_latency} of the paper validate that the predictions obtained from our latency predictor effectively align with the real-test values.

\subsection{ImageNet classification}
We use pre-trained CNN models in the official torchvision website to initialize our backbone parameters, and finetune the overall models for 100 epochs. The initial learning rate is set as 0.01$\times$batch size/128, and decays with a cosine shape. The training batch size is determined on the model size and the GPU memory. For example, we train our LAUD-ResNet-101 on 8 RTX 3090 GPUs with the batch size of 512, and the batch size for LAUD-ResNet-50 is doubled. We use the same weight decay and the standard data augmentation as in the RegNet paper \cite{radosavovic2020designing}. For our own hyper-parameter $\tau$ in Eq.~(1) of the paper, this Gumbel temperature $\tau$ exponentially decreases from 5 to 0.1 in the training procedure. For the training hyper-parameter in Eq. (2), we simply fix $\alpha=10,\beta=0.5$ and $T=4.0$ for all dynamic models. We conduct a very simple grid search with a RegNet for $\beta\in\{0.3,0.5\}$ and $T\in\{1.0,4.0\}$ to determine their values.

\subsection{COCO object detection \& instance segmentation } \label{settings_for_det_seg}
We use the standard setting suggested in \cite{lin2017feature,lin2017focal,he2017mask,zhang2023dense,cheng2022masked}, except that we decrease the learning rate for our pre-trained backbone network. We simply set a learning rate multiplier 0.5 for Faster R-CNN \cite{ren2015faster}, 0.2 for RetinaNet \cite{lin2017focal} and 0.5 for Mask R-CNN \cite{he2017mask}. \textcolor{black}{For DDQ-DETR and Mask2Former, we follow the standard setting and set the learning rate multiplier to 0.1.} As for the additional loss items, the hyper-parameters are kept the same as training our classification models, except that the temperature is fixed as 0.1 in the 12 training epochs. The input images are resized to a short side of 800 with a long side not exceeding 1333.

\section{More experimental results}
In this section, we report more experimental results which are not presented in the main paper.

\subsection{Latency prediction}\label{supp_results_latency_pred}


\noindent\textbf{Design of channel masker.} We mentioned in Sec.~\ref{sec:arch} that our channel maskers are designed as a 2-layer MLP with reduced hidden units. This design is determined under the guidance of our latency predictor. Specifically, we compare the latency of two choices with a a LAUD$^\mathrm{c}$-ResNet-101 on TX2: 1 linear layer and our 2-layer MLP. The latency numbers of the channel maskers in 4 stages and their ratios to those of ResNet blocks are reported in Table~\ref{tab_channel_masker}. The results reveal that in late stages where channel numbers are large, our 2-layer MLP with reduced hidden units significantly reduces the latency.

\noindent\textbf{Batch size.} The relationship between inference latency and batch size of LAUD-ResNet-50 on the desktop-level GPU, RTX3060, is presented in \figurename~\ref{fig:latency_bs_3060}. The phenomenon is similar to the serven-end GPUs we present in the main paper (\figureautorefname~\ref{fig:latency_bs}).

\subsection{ImageNet classification}\label{supp_results_IN_cls}
\textbf{Results on more hardware devices.} In \figurename~\ref{fig:main_results} of the paper, we report the ImageNet classification results of LAUD-ResNet on V100 and TX2, and those of LAUD-RegNet-Y on TX2 and Nano.
Here we present the results on other hardware platforms. From the results in \figurename~\ref{fig_3060_nano}, we can find that the optimal dynamic-inference paradigms can depend on the backbone and hardware devices. For example, channel skip demonstrate its advantages on Nano for LAUD-ResNet-50 and ResNet-101, while layer skipping significantly outperform the other two schemes on the more powerful devices, RTX3060 and RTX3090.

\subsection{Visualization results}\label{sec_more_vis}
Here we present more visualization results of the regions selected by our masker in the 3-rd block of a LAUD$^\mathrm{s}$-ResNet-101 ($S_{\mathrm{net}}$=4-4-2-1) in \figurename~\ref{fig_vis_supp}, which demonstrate that our spatially adaptive inference paradigm can effectively locate the most task-related areas in image features, and reduce the unnecessary computation on those background areas.

\vskip -0.2in
\begin{IEEEbiography}[{\includegraphics[width=1in,height=1.25in,clip,keepaspectratio]{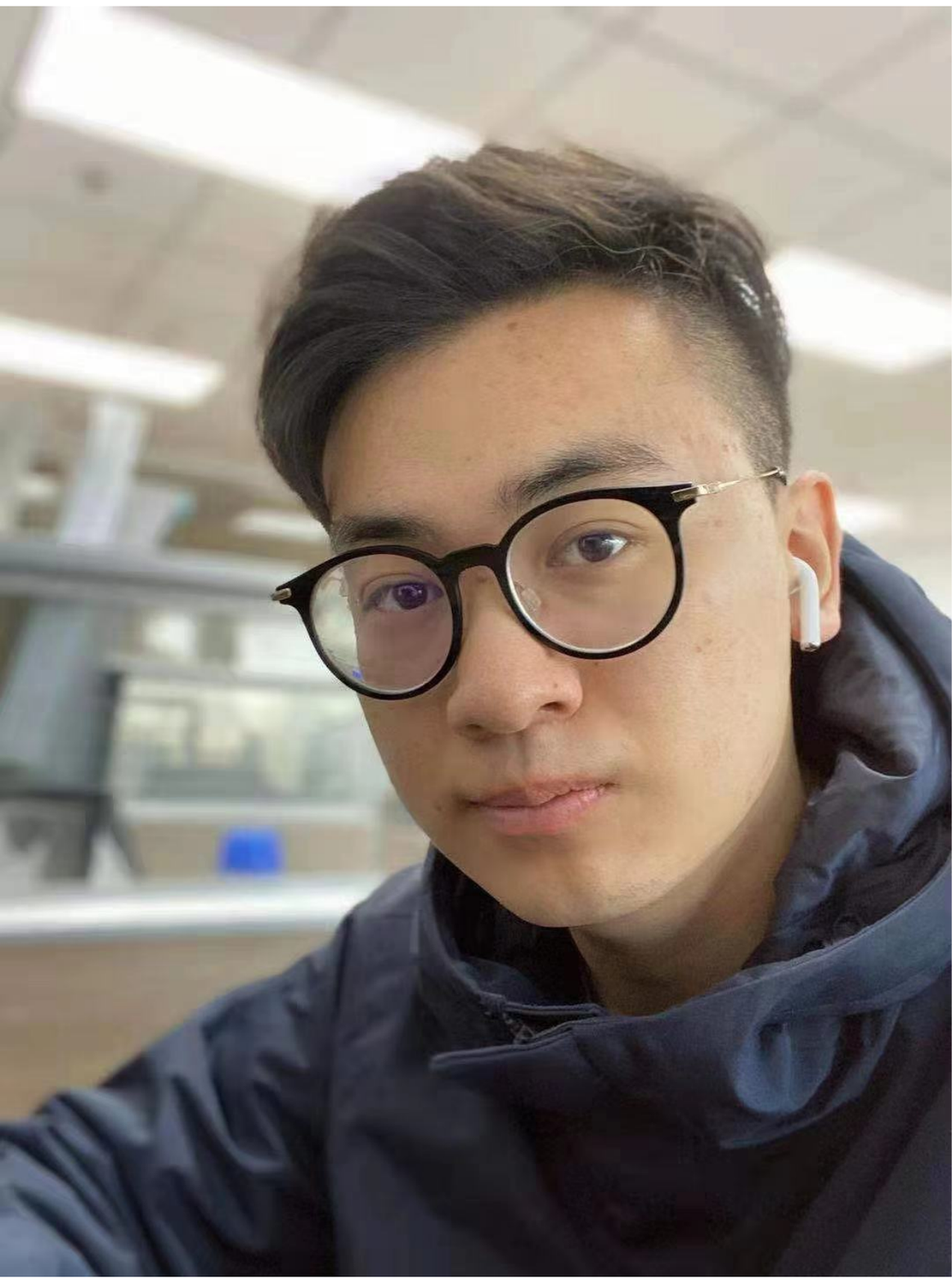}}]{Yizeng Han} received the B.S. degree from the Department of Automation, Tsinghua University, Beijing, China, in 2018. He is currently pursuing the Ph.D. degree in control science and engineering with the Department of Automation, Institute of System Integration in Tsinghua University. His current research interests include computer vision and deep learning, especially in dynamic neural networks.
\end{IEEEbiography}
\vskip -0.2in

\begin{IEEEbiography}[{\includegraphics[width=1in,height=1.25in,clip,keepaspectratio]{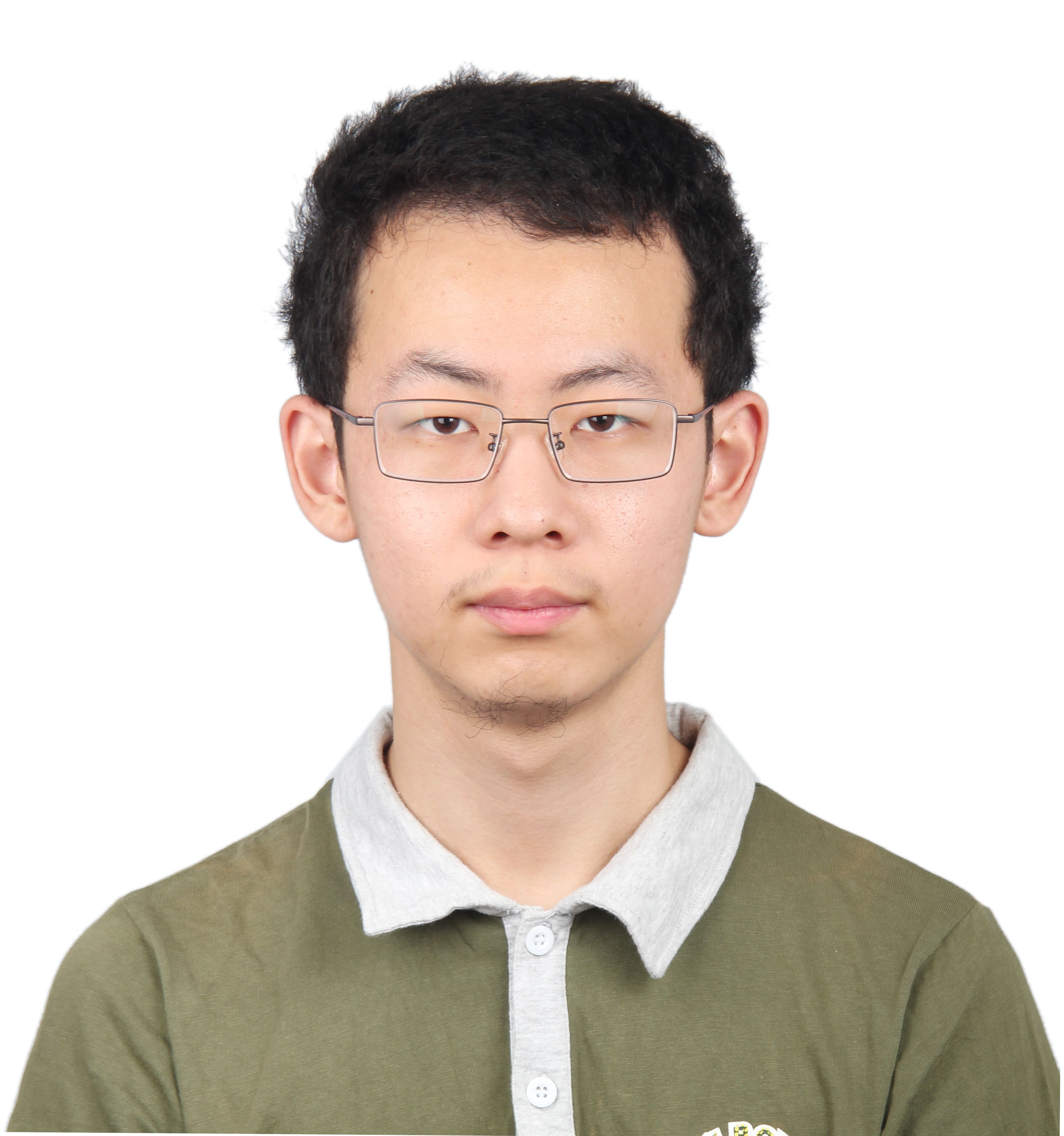}}]{Zeyu Liu} is an undergraduate student at the Department of Computer Science and Technology, Tsinghua University, Beijing, China. His current research interests include computer vision, deep learning and general AI.
\end{IEEEbiography}
\vskip -0.2in

\begin{IEEEbiography}[{\includegraphics[width=1in,height=1.25in,clip,keepaspectratio]{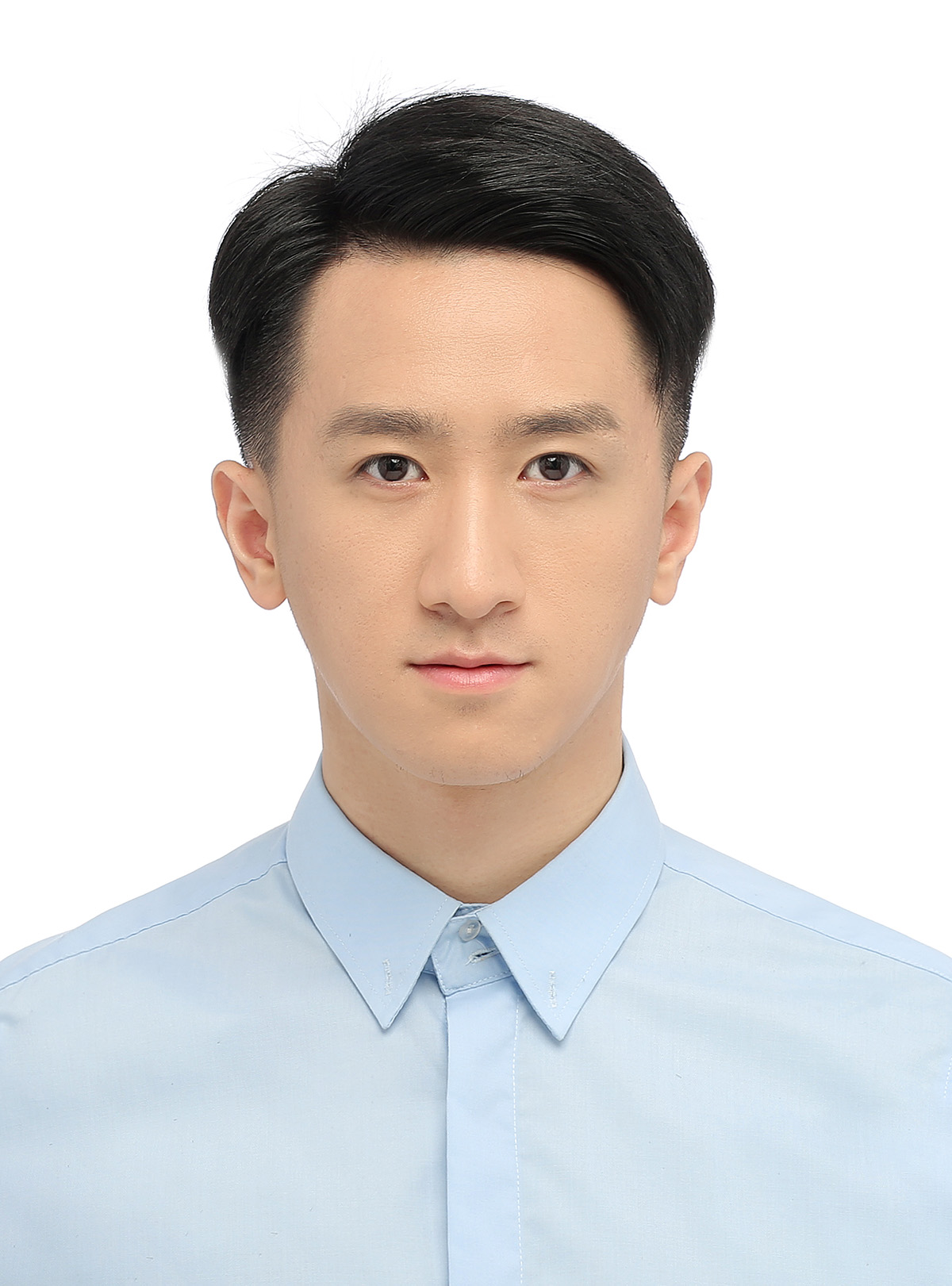}}]{Zhihang Yuan} received his Bachelor's degree in 2017 and his Ph.D. in Computer Science in 2022  from Peking University. He currently focuses on AI research, with a specific interest in the compression of neural networks, and software-hardware co-optimization. In 2021, he joined Houmo AI and has since contributed to the design of AI accelerators.
\end{IEEEbiography}
\vskip -0.2in

\begin{IEEEbiography}[{\includegraphics[width=1in,height=1.25in,clip,keepaspectratio]{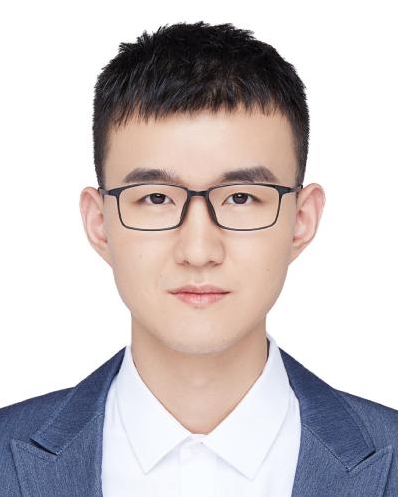}}]{Yifan Pu}
received the B.S. degree in automation from Beihang University, Beijing, China, in 2020. He is currently pursuing the M.S. degree with the Department of Automation, Tsinghhua University, Beijing, China. His research interests include computer vision, machine learning and deep learning.
\end{IEEEbiography}
\vskip -0.2in

\begin{IEEEbiography}[{\includegraphics[width=1in,height=1.25in,clip,keepaspectratio]{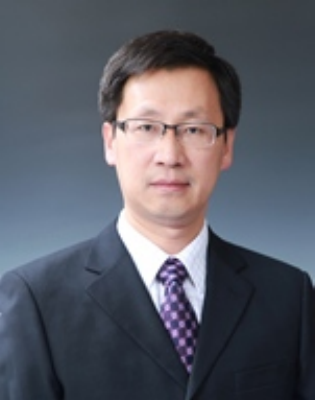}}]{Shiji Song} (SM'17) received the Ph.D. degree in mathematics from the Department of Mathematics, Harbin Institute of Technology, Harbin, China, in 1996. He is currently a Professor with the Department of Automation, Tsinghua University, Beijing, China. He has authored over 180 research papers. His current research interests include pattern recognition, system modeling, optimization and control.
\end{IEEEbiography}
\vskip -0.2in

\begin{IEEEbiography}[{\includegraphics[width=1in,height=1.25in,clip,keepaspectratio]{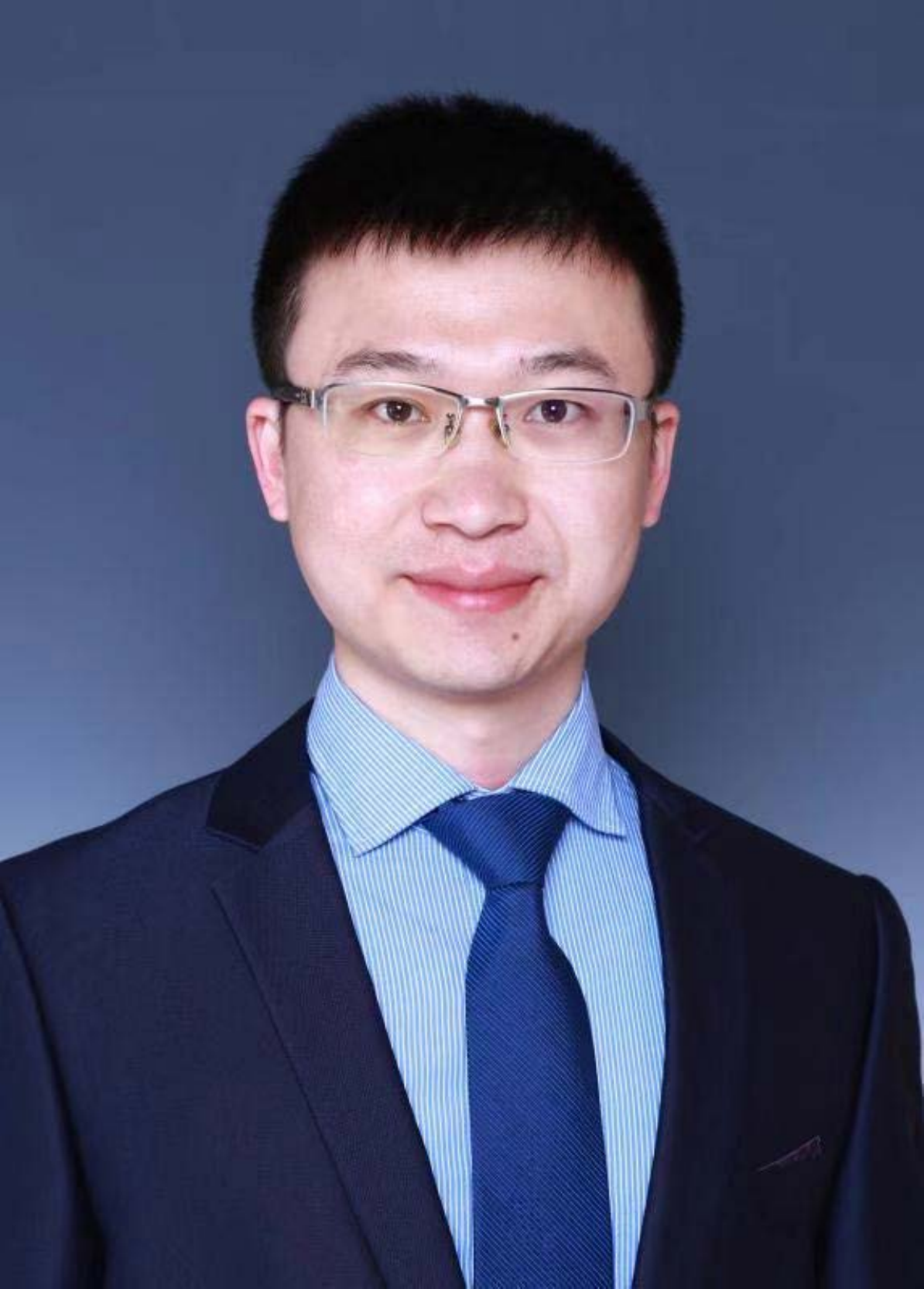}}]{Gao Huang}received the B.S. degree from the School of Automation Science and Electrical Engineering, Beihang University, in 2009, and the Ph.D. degree from the Department of Automation, Tsinghua University, in 2015. He was was a Post-Doctoral Researcher with the Department of Computer Science, Cornell University, Ithaca, USA from 2015 to 2018. He is currently an associate professor at the Department of Automation, Tsinghua University. His research interests include machine learning and computer vision.
\end{IEEEbiography}

\end{document}